\documentclass[11pt]{article}
\usepackage[preprint]{acl}

\usepackage{times}
\usepackage{latexsym}

\usepackage[T1]{fontenc}

\usepackage[utf8]{inputenc}

\usepackage{microtype}

\usepackage{inconsolata}

\usepackage{subcaption}
\usepackage{graphicx}

\usepackage{amssymb}

\usepackage{amsmath}

\usepackage{algorithm}
\usepackage{algorithmic}

\usepackage{booktabs}
\usepackage{multirow}
\usepackage{float}
\usepackage{tabularx}

\title{Softpick: No Attention Sink, No Massive Activations with Rectified Softmax}

\author{%
  Zayd M. K. Zuhri\thanks{Equal contribution} 
  \quad
  Erland Hilman Fuadi\footnotemark[1]\
  \quad
  Alham Fikri Aji \\ 
  MBZUAI\\
  \texttt{zayd.zuhri@mbzuai.ac.ae}
}

\def\softpick{\operatorname{Softpick}}
\def\softmax{\operatorname{Softmax}}
\def\relu{\operatorname{ReLU}}
\def\attention{\operatorname{Attention}}
\def\abs{\operatorname{Abs}}

\begin{document}
\maketitle
\begin{abstract}
  We introduce softpick, a rectified, not sum-to-one, drop-in replacement for softmax in transformer attention mechanisms that eliminates attention sink and massive activations. Our experiments with 340M and 1.8B parameter models demonstrate that softpick achieves 0\% sink rate consistently. The softpick transformers produce hidden states with significantly lower kurtosis and creates sparse attention maps. Quantized models using softpick outperform softmax on standard benchmarks, with a particularly pronounced advantage at lower bit precisions. Our analysis and discussion shows how softpick has the potential to open new possibilities for quantization, low-precision training, sparsity optimization, pruning, and interpretability. Our code: \url{https://github.com/zaydzuhri/softpick-attention}.
\end{abstract}

\section{Introduction}

The softmax function is widely used in statistics and particularly in machine learning as a way to normalize a vector of real numbers into a probability distribution. It has since been adopted as the de facto function to normalize the scores in the attention mechanism \citep{bahdanau2014neural, sukhbaatar2015endtoend} used in the transformer architecture \citep{vaswani2017attention}. The use of softmax in attention was a natural choice, since it represents the probability of a query "matching" to a key among many keys, which can be used to return values each weighted by that probability. However, our modern use of attention makes us question this intuition \citep{miller2023softmax1, smith2024attndiffusion}: why must they be non-zero probabilities that sum to one?

\begin{figure}[t!]
    \centering
    \includegraphics[width=\linewidth]{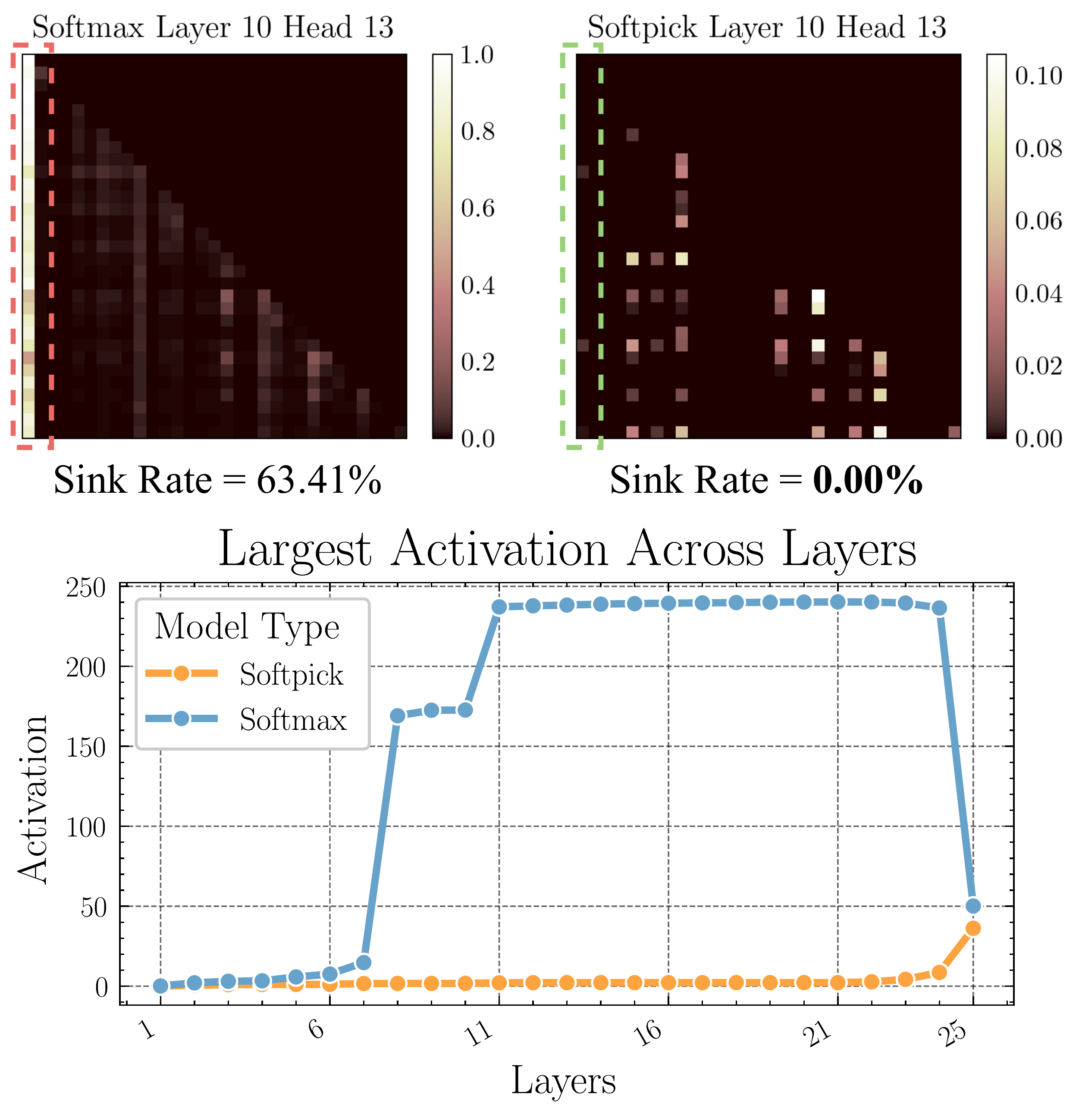}
    \caption{(Top) Comparison between the attention maps when using softmax vs softpick and overall sink rate of the 340M models. (Bottom) Largest hidden state activation per layer of the 340M models.}
    \label{fig:first-page}
\end{figure}

Softmax has great training stability with a neat Jacobian matrix resulting in a dense gradient, bounded Frobenius norm, and built-in regularization \citep{saratch2024rethinking}, and has been proven to be sufficiently expressive in its non-linearity \citep{chiang2025transformersuniformtc0}. Despite the benefits, softmax brings several oddities detrimental to language modeling. Its sum-to-one nature forces out a now well-observed behavior named attention sink \citep{xiao2023efficient} where attention heads allocate a significant score towards a specific token, often the initial BOS token, which itself is semantically irrelevant \citep{gu2024attention}. These sinks are, as far as we know, harmless to downstream performance. However, another symptom of this same effect is the appearance of extreme hidden state activations, dubbed massive activations \citep{sun2024massive}, which grow larger as the model scales \citep{bondarenko2023quantizable}. These activations are especially problematic for quantization, with entire algorithms built around them. Additionally, most low-precision training approaches try to work around these massive activations as they are unwieldy in low-bit settings. Removing these massive activations would open up more possibilities.

Even with these quirks, softmax remains widely used in attention due to its stability and effectiveness. In this paper, we propose the softpick function in an attempt to find a normalization function more fitting for attention and its use in transformers. We mitigate both attention sink and massive activations by reshaping the softmax function to avoid strict sum-to-one behavior and allowing rectified outputs. Our contributions are as follows:
\begin{enumerate}
    \item We propose the softpick function as a drop-in replacement to softmax in attention.
    \item We experiment by training 340M and 1.8B parameter transformer models from scratch and show how softpick compares to softmax in benchmarks and training behavior. We also show that quantized softpick models down to 2-bit precision outperform softmax.
    \item We analyze the resulting models and show that softpick returns more legible attention maps, a sink rate of 0\%, and no massive activations in the hidden states.
    \item We investigate the subpar performance of softpick at the 1.8B scale and provide two possible hypotheses that extend our understanding of softmax-like functions for attention.
\end{enumerate}

\begin{figure*}[t]
    \centering
    \includegraphics[width=0.9\textwidth]{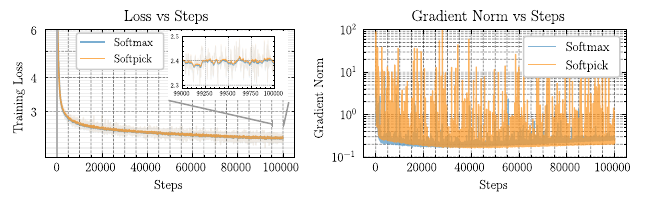}
    \caption{Training loss and gradient norm during training of 340M models.}
    \label{fig:main-loss}
\end{figure*}

\section{Background}

Attention sink was characterized by \citet{xiao2023efficient} as heads that allocate disproportionate attention to semantically weak but frequent tokens, most commonly the BOS token. \citet{barbero2025llms} argue that such sink-like heads can mitigate over-mixing by acting as approximate no-ops, while \citet{gu2024attention} systematically study correlates of the phenomenon and report that sinks appear broadly in sufficiently trained transformer language models. Together, these results suggest that sink behavior is not an artifact of a specific setup, but is closely tied to attention normalization, in particular the sum-to-one constraint imposed by softmax.

Massive activations, identified by \citet{sun2024massive} as rare but extremely large hidden-state values, pose a challenge for quantization and low-precision training \citep{dettmers2022llmint88bitmatrixmultiplication}. \citet{sun2024massive} further link these outliers to self-attention and softmax-driven sink behavior, motivating mitigation strategies that relax strict normalization. For example, \citet{gu2024attention} show that removing normalization reduces sinks, but sinks reappear once normalization is reintroduced. KV bias variants similarly require additional learned parameters and can degrade with depth \citep{gu2024attention,sun2024massive}. Softmax+1 relaxations add a constant to the denominator \citep{miller2023softmax1} and are evaluated in \citet{kaul2024attention}, while \citet{openai2025gptoss120bgptoss20bmodel} adopt a learnable-bias variant at scale to relax strict sum-to-one behavior. However, \citet{owen2025refined} find KV bias techniques to be unreliable for mitigating massive activations, and architectural alternatives such as gated attention require additional parameters \citep{qiu2025gated}.  We propose softpick as a drop-in alternative that mitigates both sinks and massive activations without adding parameters or custom optimizers.

\section{Method}

\subsection{Softpick Function}
Given a vector $\mathbf{x} \in \mathbb{R}^N$, we define the softpick function as
\begin{equation}
    \softpick(\mathbf{x})_i = \frac{\relu(e^{x_i} - 1)}{\sum_{j=1}^{N}|e^{x_j} - 1|}
\end{equation}
with the rectified linear unit defined as $\relu(x) = max(x, 0)$ and the absolute value function written as $|x|$. Similar to softmax, we need a numerically safe version to use in practice. We define numerically safe softpick as
\begin{equation}
    \softpick(\mathbf{x})_i = \frac{\relu(e^{x_i - m} - e^{-m})}{\sum_{j=1}^{N}|e^{x_j-m} - e^{-m}| + \epsilon}
\end{equation}
where $m$ is the maximum value inside $\mathbf{x}$ and $0<\epsilon \ll 1$ is to avoid division by zero in the case that all inputs are exactly zero. This function is a drop-in replacement to softmax in the attention mechanism:
\begin{equation}
    \attention(\mathbf{Q}, \mathbf{K}, \mathbf{V})=\softpick\left(\frac{\mathbf{Q}\mathbf{K}^T}{\sqrt{d_k}}\right)\mathbf{V}
\end{equation}

\subsection{Design Rationale}
To understand the reasoning behind each component of the softpick function, it is best we go through a step-by-step recreation of the formula. We start from the vanilla softmax function.
\begin{equation}
    \softmax(\mathbf{x})_i = \frac{e^{x_i}}{\sum_{j=1}^{N}e^{x_j}}
\end{equation}
The reason why we still do not modify this function too much is that the characteristics of softmax in training are ideal for attention. We want to maintain as much of the Jacobian matrix as possible and keep the gradient norm bounded that way. The next step would be to induce "null attention", or allowing attention to output zeros. We can achieve this by moving down the exponential by one, making it output negative values $-1<y<0$ for negative inputs. We then rectify this output with ReLU.
\begin{equation}
    \softmax(\mathbf{x})_i = \frac{\textcolor{orange}{\relu(}e^{x_i}\textcolor{orange}{-1)}}{\sum_{j=1}^{N}\textcolor{orange}{\relu(}e^{x_j}\textcolor{orange}{-1)}}
\end{equation}
This function has several desirable properties. First, it allows zero-valued outputs, which potentially enables sparsity optimizations and minimizes noise from the accumulation of unneeded values. Second, unlike softmax, the denominator allows for zero values which might reduce the scaling down of attention scores at long sequence lengths. This allows for a sharper attention map, where only the positive inputs receive attention scores that sum to one. However, this is also its weakness. We experimented with this rectified-only softmax and found that it diverges from baseline softmax after a longer training period, and hypothesize that this is caused by heads "dying" and not being able to recover from outputting only zeros. This is obvious when we look at the Jacobian of the function, where negative inputs do not receive any gradients because they do not contribute to the output nor the denominator. Moreover, using this modification does not result in getting rid of attention sink. We go into more detail on this and show the experimental results of using a similar function in Appendix~\ref{sec:other-experiments-appendix}. To fix the issue of negative inputs receiving minimal gradients, we allow them to contribute to the denominator.
\begin{equation}
    \softmax(\mathbf{x})_i = \frac{\relu(e^{x_i}-1)}{\sum_{j=1}^{N}\textcolor{orange}{\abs(}e^{x_j}-1\textcolor{orange}{)}}
\end{equation}
The absolute function $\abs(x)$ or $|x|$ allows us to rectify each element for summation, resulting in a positive sum, but does not alter the derivative of $e^x$ other than its sign, i.e. $\frac{d}{dx}|e^x|=-e^x$ for $x<0$. This is crucial to one of our goals, which is to maintain the desired properties from the Jacobian of softmax. This way, gradients can flow even when the input to the function is negative. More importantly, this asymmetry between the numerator and denominator removes the strict requirement to sum to one, which is the main cause of attention sink.

\begin{table*}[t]
\centering
\small
\setlength{\tabcolsep}{4pt}
\begin{tabular}{l|l|rrr|rrr}
\toprule
\multirow{2}{*}{Task} & \multirow{2}{*}{Metric} &
\multicolumn{3}{c|}{340M} & \multicolumn{3}{c}{1.8B} \\
\cmidrule(lr){3-5}\cmidrule(lr){6-8}
 &  & Softmax & Softpick & $\Delta$ & Softmax & Softpick & $\Delta$ \\
\midrule
\multirow{2}{*}{Arc Easy} & Acc Norm \hfill $\uparrow$ & 56.61 & 56.73 & \textcolor{green!60!black}{+0.12} & 67.21 & 62.04 & \textcolor{red!70!black}{-5.17}\\
\cmidrule(lr){2-8}
 & Acc \hfill $\uparrow$ & 60.35 & 61.11 & \textcolor{green!60!black}{+0.76} & 72.73 & 68.60 & \textcolor{red!70!black}{-4.13}\\
\midrule
\multirow{2}{*}{Lambada} & Acc \hfill $\uparrow$ & 36.25 & 36.21 & \textcolor{red!70!black}{-0.04} & 49.43 & 43.51 & \textcolor{red!70!black}{-5.92}\\
\cmidrule(lr){2-8}
 & Perplexity \hfill $\downarrow$ & 30.33 & 28.63 & \textcolor{green!60!black}{-1.70} & 11.38 & 15.81 & \textcolor{red!70!black}{+4.43}\\
\midrule
\multirow{2}{*}{Piqa} & Acc Norm \hfill $\uparrow$ & 66.59 & 66.49 & \textcolor{red!70!black}{-0.10} & 73.61 & 70.89 & \textcolor{red!70!black}{-2.72}\\
\cmidrule(lr){2-8}
 & Acc \hfill $\uparrow$ & 66.97 & 66.59 & \textcolor{red!70!black}{-0.38} & 73.78 & 71.27 & \textcolor{red!70!black}{-2.51}\\
\midrule
\multirow{2}{*}{Sciq} & Acc Norm \hfill $\uparrow$ & 74.90 & 77.30 & \textcolor{green!60!black}{+2.40} & 86.40 & 80.40 & \textcolor{red!70!black}{-6.00}\\
\cmidrule(lr){2-8}
 & Acc \hfill $\uparrow$ & 83.20 & 83.60 & \textcolor{green!60!black}{+0.40} & 90.10 & 87.00 & \textcolor{red!70!black}{-3.10}\\
\midrule
Wikitext & Word Perplexity $\downarrow$ & 23.85 & 24.32 & \textcolor{red!70!black}{+0.47} & 15.10 & 17.87 & \textcolor{red!70!black}{+2.77}\\
\bottomrule
\end{tabular}
\caption{Comparison of softpick vs softmax on downstream tasks for 340M and 1.8B models. $\uparrow$=Higher is better, $\downarrow$=Lower is better. $\Delta$ = Softpick - Softmax.}
\label{tab:two_column_spanning_benchmark}
\end{table*}

\subsection{Derivative}
Given $\mathbf{s} = \softpick(\mathbf{x}) \in \mathbb{R}^N$ as the output, the partial derivative or elements of the Jacobian matrix of the (numerically safe) softpick function can be written as such:
\begin{equation}
    \frac{\partial s_i}{\partial x_j} = \frac{e^{x_j-m}}{\Sigma} \left( \delta_{ij} \operatorname{step}(x_i) - \operatorname{sign}(x_j) s_i \right)
\end{equation}
where $\delta_{ij}$ is the Kronecker delta, equal to $1$ if $i=j$ and $0$ otherwise, with $\operatorname{step(x)}$ being the step function that returns $0$ if $x \leq 0$ and $1$ if $x > 0$ and $sign(x)$ being the sign function that returns $-1$ if $x < 0$ and $1$ if $x \geq 0$. Additionally, $\Sigma = \sum_{k=1}^{N}|e^{x_k-m} - e^{-m}| + \epsilon$ is the denominator of the softpick function. Although unlike softmax where only the outputs need to be kept for a naive backward pass implementation, the softpick derivative is nonetheless trivial for implementations that recompute the inputs, as is done in FlashAttention.

\subsection{FlashAttention}
The FlashAttention algorithm \citep{dao2022flashattentionfastmemoryefficientexact, dao2023flashattention2} allows for online single-pass computation of attention, bypassing the quadratic memory requirement. Because the $\relu(x)$ and absolute $|x|$ functions uphold the multiplicative property for positive-valued multipliers (such as $e^x$ for any $x$), deriving an online version of softpick is possible, thus making it compatible with FlashAttention. We provide the algorithms for the more recent FlashAttention-2 version of the forward pass and backward pass that use the softpick function instead of softmax in Appendix~\ref{sec:flashattention-appendix}.

\section{Experiments}
\label{sec:experiments}
To validate the viability of the softpick function over softmax, we conduct experiments on 
Llama-style (pre-norm, RoPE, SwiGLU MLP) transformers trained from scratch. We train 4 models, two for each size: 340M and 1.8B parameters, each with a softmax variant and a softpick variant. The detailed training configuration for all models is explained in Appendix~\ref{sec:training-configuration} shown in Table~\ref{tab:training-config}. We use the flash-linear-attention repository \citep{yang2024fla} and their Flame training framework (built on top of torchtitan \citep{liang2025torchtitan}) because of their easily modifiable and fast triton \citep{tillettriton} kernels. We train each 340M model on a total of 52 billion tokens and each 1.8B model on 104 billion tokens sampled from the 100B subset of the fineweb-edu \citep{lozhkov2024fineweb-edu} dataset. Each training run took approximately 18 hours for the 340M models and 116 hours for the 1.8B models on 8xH100 GPUs.

We also execute short training runs of models for analysis and comparison against other sink-mitigation / sink-free attention variants. Unless specified otherwise, we train these models on 2$\times$A100 GPUs or 4$\times$MI210 AMD GPUs. Each model uses the 340M configuration in Appendix~\ref{sec:training-configuration} on Table~\ref{tab:training-config}. We intentionally limit these runs to 10k steps because attention sinks in the vanilla softmax baseline already emerge within this window, making it sufficient for diagnosing sink behavior and activation outliers while keeping the ablation compute manageable. As these models are minimally trained, we do not run benchmarks with them.

\begin{figure*}[t]
    \centering
    \includegraphics[width=\linewidth]{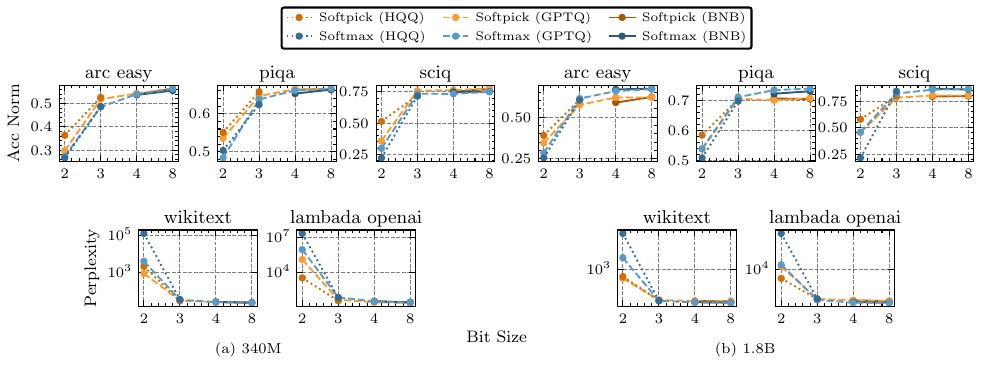}
    \caption{Quantization results of softmax vs.\ softpick across model scales. We deliver results on 2, 3, 4, and 8-bit quantization. Full tabular results in Appendix~\ref{sec:quantization-appendix} and \ref{sec:quantization-appendix-1.8B}.}
    \label{fig:scores_vs_all_qbit}
\end{figure*}

\section{Results}
\paragraph{Training Loss and Gradients}
We present the resulting training loss and gradient norm of the 340M model over 100,000 training steps (52B tokens) in Figure~\ref{fig:main-loss}. We observed that the training loss of the softpick transformer tracks very closely to the vanilla softmax transformer, with only a 0.004 gap in the final training loss. This supports our design decision to maintain as much of the important components of the softmax function as possible. However, this does not hold at scale, as we observe a larger gap of 0.12 in the training loss of the larger 1.8B model. This will be reflected in the benchmark results. Tangentially, we observe that the total gradient norm of the models' parameters during training exhibits noteworthy behavior. At the start up to a third of the way into training, the gradient norm of the softpick model is higher than softmax, but goes down and stays below softmax until the end, and does not rise at as fast of a rate as softmax. Also important to note is that the magnitude of some gradient norms of the softpick model are much higher than the gradient norm peaks of softmax. We use gradient clipping with a maximum norm of 1.0 and do not see any instability caused by these large gradients.

\paragraph{Benchmarks}
We benchmark all models with general downstream tasks that are appropriate for the model sizes. We use the LM Evaluation Harness \citep{eval-harness} to evaluate on these benchmarks: AI2 Reasoning Challenge (ARC-e) \citep{clark2018arc}, Lambada \citep{paperno-etal-2016-lambada}, specifically the OpenAI variant, PIQA \citep{bisk2019piqa}, and SciQ \citep{welbl2017sciq}. All 4 benchmarks report accuracy, with an addition of perplexity for Lambada. We also measure validation perplexity on Wikitext \citep{merity2016pointer}. See Table~\ref{tab:two_column_spanning_benchmark} for full results. We observe equal if not slightly better performance from softpick compared to softmax on the 340M model. However, the 1.8B models show that softpick does not seem to scale well, at least when using the same hyperparameters as softmax. Overall accuracy on benchmarks and perplexity are worse at this scale.

\paragraph{Quantization}
We also benchmark the effect of quantization on the trained models. We verified this using HQQ \citep{HQQ_2023}, BitsandBytes (BNB) \citep{dettmers2022llmint88bitmatrixmultiplication, dettmers2023qloraefficientfinetuningquantized}, and GPTQ \citep{frantar2023gptqaccurateposttrainingquantization}. These methods encompass different approaches, including those that utilize calibration data and those that do not. We use the same benchmarks as before. The results can be seen in Figure~\ref{fig:scores_vs_all_qbit} with more complete tables in Appendix~\ref{sec:quantization-appendix}. At the 340M scale, when using softpick, the quantized models are consistently better than when softmax is used, and quantization is less damaging as the precision goes down. This becomes less clear at the 1.8B as the gap in performance of the base models is larger. However, the gap does close at lower precisions.

\section{Analysis}
\begin{table}[t]
\small
\centering
\setlength{\tabcolsep}{4pt}
\begin{tabular}{llcc}
\toprule
Size & Model & $\epsilon_s{=}0.2 $ & $\epsilon_s{=}0.3$ \\
\midrule
\multirow{2}{*}{340M} & Softmax   & 68.28 & 63.41 \\
                      & Softpick  & 0.00  & 0.00  \\
\midrule
\multirow{2}{*}{1.8B} & Softmax   & 41.73 & 14.96 \\
                      & Softpick  & 0.00  & 0.00  \\
\bottomrule
\end{tabular}

\vspace{0.5em}

\begin{tabular}{llcccc}
\toprule
Size & Model & Kurt. & Min. & Max. & Spars.\% \\
\midrule
\multirow{2}{*}{340M} & Softmax   & 33510.81 & -207.55 & 240.27 & 4.53* \\
                      & Softpick  &   340.96 &  -45.03 &  36.21 & 99.34 \\
\midrule
\multirow{2}{*}{1.8B} & Softmax   & 74456.81 & -173.48 & 371.43 & 4.28* \\
                      & Softpick  &  2193.27 &  -51.05 & 101.51 & 95.63 \\
\bottomrule
\end{tabular}
\caption{Sink rate, activation statistics, and attention sparsity for softmax vs. softpick. Sparsity is the \% of exact zeros in the causal (lower-triangular) attention matrix; for softmax, zeros arise only from numerical underflow.}
\label{tab:analysis-table}
\vspace{2pt}
\end{table}

\begin{figure}
\centering
\begin{subfigure}{0.4\textwidth}
    \includegraphics[width=\linewidth]{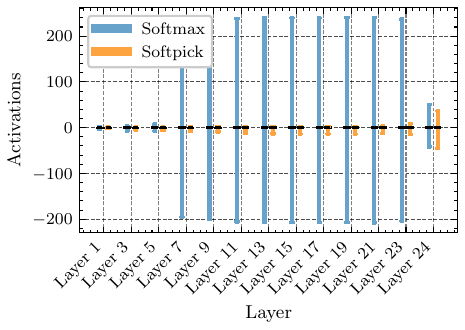}
    \caption{340M models}
\end{subfigure}
\begin{subfigure}{0.4\textwidth}
    \includegraphics[width=\linewidth]{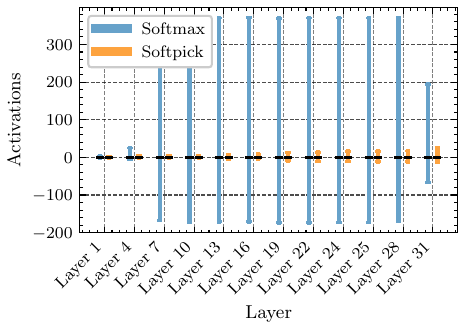}
    \caption{1.8B models}
\end{subfigure}
\caption{Box plots of the hidden state activations at some layers of the softmax (left,
blue) and softpick (right, orange) models.}
\label{fig:boxplots}
\end{figure}

\begin{table}[t]
\small
\centering
\setlength{\tabcolsep}{4pt}
\begin{tabular}{lcc}
\toprule
Model & $\epsilon_s{=}0.2$ & $\epsilon_s{=}0.3$ \\
\midrule
Softmax & 47.97 & 34.89 \\
Softpick & 0.02 & 0.00 \\
Gated Attention & 5.00 & 2.00 \\
GPT-OSS Sink & 19.34 & 8.78 \\
Rectified-Only & 33.08 & 25.60 \\
ReLU Softmax & 8.16 & 4.56 \\
Scaled Softpick & 0.01 & 0.00 \\
Softmax+1 & 24.38 & 16.36 \\
\bottomrule
\end{tabular}

\vspace{0.5em}

\begin{tabular}{lcccc}
\toprule
Model & Kurt. & Min. & Max. & Spars.\% \\
\midrule
Softmax & 6452.72 & -270.50 & 178.88 & 0.16* \\
Softpick & 281.57 & -59.06 & 25.36 & 92.74* \\
Gated Attention & 135.62 & -49.09 & 23.31 & 0.73* \\
GPT-OSS Sink & 1800.34 & -36.25 & 106.62 & 0.05* \\
Rectified-Only & 11054.50 & -312.25 & 271.00 & 60.70* \\
ReLU Softmax & 452.59 & -61.28 & 60.62 & 29.08* \\
Scaled Softpick & 170.70 & -53.69 & 30.81 & 91.95* \\
Softmax+1 & 574.16 & -47.56 & 83.88 & 0.12* \\
\bottomrule
\end{tabular}
\caption{Sink rate, activation statistics, and attention sparsity for sink-free attention variants. Sparsity is the \% of exact zeros in the causal (lower-triangular) attention matrix; for softmax, zeros arise only from numerical underflow.}
\label{tab:ablations_models_metrics}
\vspace{2pt}
\end{table}

\subsection{Attention Maps}
\label{sec:attention-map}
\begin{figure*}
    \centering
    \small
    \includegraphics[width=0.95\linewidth]{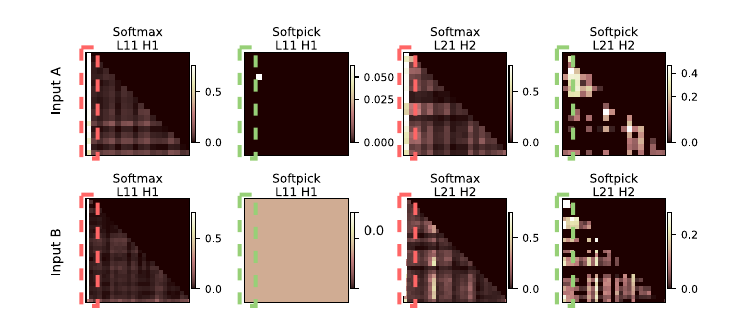}
    \caption{Attention maps of softmax and softpick 340M models on 2 different input texts. Two heads are visualized: Head 1 of Layer 11 and Head 2 of Layer 21. See more attention maps in Appendix~\ref{sec:attention-maps-appendix}.}
    \label{fig:attention_maps}
\end{figure*}
To understand the difference that the softpick function makes compared to softmax, it is best to observe the attention maps directly. Attention maps are the matrices resulting from the computation of $\mathbf{A} = \operatorname{Soft\{max,pick\}}(\mathbf{QK}^T/\sqrt{d_k})$. We prepare three text inputs as samples to run through the models and extract the attention maps. We provide some clear example attention maps from the 340M models in Figure~\ref{fig:attention_maps} and show more attention maps of all models in Appendix~\ref{sec:attention-maps-appendix}.

First, it is clear that softpick behaves very differently in regards to how it normalizes attention scores compared to softmax. Due to the rectified numerator in softpick, scores are functionally sparse and concentrated into specific spots and regions, surrounded by actual zero scores. Although more general heads that attend to a wider range of tokens do still exist in the softpick model. Second, and more important to our thesis, is that attention sinks are nowhere to be seen with softpick.

To see the sparsity effect of softpick, we calculate the sparsity ratio of the attention map for softpick and softmax by running 10 text samples through the model, collecting the attention maps, and averaging the the sparsity, which is calculated by counting the number of exact zeros. Note that since this is a causal language model, we only consider the lower triangular scores of the attention map. As shown in Table \ref{tab:analysis-table}, our method results in attention scores with 99.34\% sparsity on the 340M model and 95.63\% on the 1.8B model. While softmax in theory cannot return exact zeros, we observe a non-zero sparsity because in practice floating point scores can underflow to zero.

\subsection{Attention Sink}

 Attention sink can be identified by the large scores on the first column in the attention map, which manifests as a bright vertical line leftmost on the attention heatmaps. Notably, the heads that have heavy attention sinks i.e. large scores on the BOS token and close to none on other tokens with softmax, become either specifically tuned to only score specific trigger tokens, or are completely shut off when not needed. An example of this can be seen in Figure~\ref{fig:attention_maps}, on head 4 of the 10th layer. The softmax model has a clear attention sink and barely attends to other tokens, while the softpick model completely shuts off the head (all zero scores) on the first input, and attends to some specific tokens on the second and third input sample text. We believe this behavior is a more concrete version of the active-dormant heads in vanilla attention, a previously observed phenomenon \citep{guo2024activedormant}. We discuss this and its implications in \S\ref{sec:discuss}.

 In addition to visually observing attention sinks, we can use the sink rate metric proposed in \citep{gu2024attention}, defined as $\frac{1}{L}\sum_{l=1}^{L}\frac{1}{H}\sum_{h=1}^{H}\mathbb{I}(\alpha_1^{l,h} > \epsilon_s)$ which is the percentage of heads in the transformer that on average has an attention score above some threshold $\epsilon_s$ in the first token, where $\alpha_1^{l,h}$ is the average value of the first column in $\mathbf{A}$ at layer $l$ and head $h$. We calculate the sink rate of both models on 1000 samples from the validation set of SlimPajama \citep{cerebras2023slimpajama} with $\epsilon_s = 0.3$ (as suggested in \citep{gu2024attention}) and also $\epsilon_s = 0.2$ and present the results in Table~\ref{tab:analysis-table}. The model using softpick has a sink rate of 0\% for both thresholds and both model sizes, which means that no attention heads are overscoring the BOS token in any scenario, effectively eliminating attention sink completely. The effects of this can be seen when we analyze the hidden activations between transformer layers and the large values inside them.

\subsection{Massive Activations}
\label{sec:outlier-analysis}
We analyze the effect of softpick on massive activations by examining the hidden state outputs after every transformer layer. We present some key metrics in Table~\ref{tab:analysis-table} and box plots of the hidden state activation distribution of every some layers of the 340M and 1.8B models in Figure~\ref{fig:boxplots}. These metrics are calculated by running 10 samples of text through the model and collecting all hidden state vectors to evaluate them all at once. The box plots visualize how large the massive activations of the softmax model are, making the boxes (Q1, median, and Q3) barely visible due to the large minimum and maximum values. This is especially prominent from the middle layers up to second to last layer. Meanwhile, the softpick models do not exhibit these massive activations. There is an exception in the last layer where it matches the softmax model, most likely due to the need to project back into the vocabulary space. This large difference can be seen clearly in the metrics as well. Kurtosis of all hidden state activations in the 340M model is significantly reduced from 33,510 to 340, a hundred-fold reduction. The minimum and maximum values are also reduced by an order of magnitude. Note that the minimum and maximum values of the softpick model are in fact from the last layer, all previous layers have even smaller activations.

\subsection{Comparison with Other Sink-Free Methods}
We compare softpick to several other sink-mitigation/sink-free attention variants using the same outlier diagnostics: sink rate (lower is better), activation statistics (kurtosis and extrema), and the fraction of exact zeros in the causal attention matrix (``Spars.\%''). We omit downstream benchmarks because these ablation runs are too short to yield meaningful capabilities (see \S\ref{sec:experiments}). Table~\ref{tab:ablations_models_metrics} reports results for $\epsilon_s \in \{0.2, 0.3\}$.

Across both thresholds, softpick (and scaled softpick) most strongly suppresses sinks, achieving near-zero sink rates while substantially reducing activation extremes relative to softmax. Other methods help but less consistently: gated attention reaches single-digit sink rates but remains dense; softmax+1 roughly halves sink rate with near-zero sparsity; rectified is more sink-prone and shows the heaviest-tailed activations despite moderate sparsity. Softmax-like methods are effectively non-sparse (aside from underflow), whereas softpick yields genuinely sparse attention with many exact zeros, consistent with its mechanism.



\section{Scalability}
\subsection{Long Context and Underscoring}
\label{sec:long_context}
\begin{table}[t]
  \centering
  \footnotesize
  \setlength{\tabcolsep}{3.5pt} 
  \renewcommand{\arraystretch}{1.05}
  \begin{tabularx}{\columnwidth}{l *{5}{>{\centering\arraybackslash}X}}
    \toprule
    Model & \multicolumn{5}{c}{Sequence Length} \\
    \cmidrule(lr){2-6}
    & 890 & 1983 & 3075 & 4167 & 4986 \\
    \midrule
    Softmax & 95.5 & 97.0 & 73.0 & 70.5 & 0.0 \\
    Softpick & 94.0 & 91.0 & 75.0 & 66.5 & 0.0 \\
    Rectified-Only & 95.5 & 68.0 & 57.5 & 24.5 & 0.0 \\
    Scalable-Softmax & 98.0 & 96.9 & 87.0 & 55.0 & 0.0 \\
    \bottomrule
  \end{tabularx}
  \caption{Passkey retrieval results including the other experiments.}
  \label{tab:other-passkey-table}
\end{table}

Unlike softmax where every value must have a non-zero score, softpick can assign zero to unneeded values, theoretically leading to better retrieval for long context use. However, we have yet to see this benefit in our testing. On the passkey retrieval task \citep{lu2024longheads}, softpick performs comparably to, but does not outperform, softmax across multiple sequence lengths. Table~\ref{tab:other-passkey-table} reports passkey retrieval accuracy across increasing sequence lengths. Performance generally degrades as context grows, reflecting the increased difficulty of retrieving the key from longer inputs. All methods collapse to $0.0$ at length 4986 because our models were trained with a maximum context length of 4096, so this setting is out-of-distribution and exceeds the trained positional range.

We hypothesize that this stems from an underscoring effect: as context length grows and attention becomes increasingly sparse, the normalization in softpick can significantly reduce the magnitude of scores assigned to the few relevant tokens, especially when many irrelevant tokens receive negative scores. This results in weaker value signals, as evidenced by the reduced scale in retrieval-specific heads that only attend to singular tokens (Appendix~\ref{sec:attention-maps-appendix}, see the color bar value of the heatmaps). We also hypothesize that this is one of the reasons why the larger 1.8B model with softpick does not perform as well as softmax. We explored scaling the query-key scores prior to softpick, inspired by Scalable-Softmax \citep{nakanishi2025scalablesoftmax} and its use in Llama~4 \citep{llama4}, but found no improvement in retrieval (Table~\ref{tab:other-passkey-table}) and observed worse training and downstream performance (Appendix~\ref{sec:other-experiments-appendix}).


\subsection{Dead Attention Heads}
\label{sec:dead_attention_heads}
\begin{figure}
    \centering
    \includegraphics[width=0.8\linewidth]{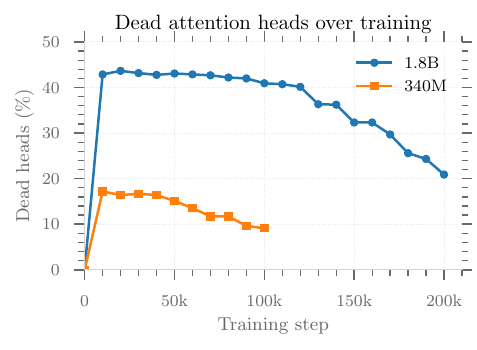}
    \caption{Percentage of dead heads on 340M and 1.8B softpick model across training steps}
    \label{fig:dead_head_percentage}
\end{figure}

We further investigate why softpick may scale poorly to larger models. Another hypothesis is that some attention heads remain dormant or dead for long periods during training, which may hinder gradient flow or reduce effective capacity. To test this, we run in-distribution inference on 5M tokens using checkpoints from 0 to 100k steps (every 10k). We label a head as dead if its maximum absolute output is close to zero ($\epsilon = 10^{-6}$) for a token and this holds for at least $95\%$ of tokens. 

From Figure~\ref{fig:dead_head_percentage}, we can see that at the start of training, the weights are randomly initialized; therefore, dead heads have not yet emerged. However, after only the first 10k steps, we observe that the number of dead heads reaches its peak. Subsequently, this count consistently decreases until the end of training. The 340M model goes from $17.19\%$ to $9.11\%$ dead heads during training, while the 1.8B model decreases from $42.87\%$ to $20.90\%$. This suggests larger models exhibit more dead heads, which may reduce capacity and/or gradient flow at scale. Since our training is limited to 100B tokens, the continued decline in dead head percentage also suggests we may be undertraining, and scaling to trillion-token regimes may improve softpick’s scalability. It is notable however that when the threshold is set to $\epsilon = 10^{-2}$, the 340M model still exhibits $22.66\%$ dead heads while the 1.8B model contains $50.39\%$ dead heads at the end of training.

We also provide downstream evaluation when we actually zero out heads with output $\epsilon <0.01$ and $\epsilon < 1e^{-6}$ to support our arguments. As we see in Table~\ref{tab:downstream_zero}, masking out these dead attention heads at inference time damages the smaller 340M model much more than the larger 1.8B model. However, for both models, the difference in performance going from masking at $10^{-6}$ to $10^{-2}$ is nearly negligible, which supports the argument for setting a higher threshold for dead attention heads.

\begin{table}[t]
\centering
\small
\setlength{\tabcolsep}{4pt}
\renewcommand{\arraystretch}{1.08}
\begin{tabular}{lcccc}
\toprule
Model & $\epsilon$ & Wikitext PPL & \shortstack{LAMBADA\\Accuracy} & \shortstack{LAMBADA\\PPL} \\
\midrule
\multirow{3}{*}{\textbf{340M}}
& -         & 26.47 & 37.01\% & 27.94 \\
& $10^{-6}$ & 65.42 & 19.02\%          & 223.02 \\
& $10^{-2}$ & 79.07 & 18.88\%          & 199.86 \\
\midrule
\multirow{3}{*}{\textbf{1.8B}}
& -         & 20.94 & 44.48\% & 14.85 \\
& $10^{-6}$ & 21.47 & 44.46\%           & 14.77 \\
& $10^{-2}$ & 21.43 & 43.68\%           & 15.11 \\
\bottomrule
\end{tabular}
\caption{Downstream performance under different dead head masking thresholds $\epsilon$.}
\label{tab:downstream_zero}
\end{table}

\section{Discussion and Implications}
\label{sec:discuss}
Across quantization, low-precision training, and efficiency considerations, a central challenge in modern transformers is the presence of massive activation outliers and attention sinks. Prior work has shown that these outliers distort scaling in activation quantization \citep{dettmers2022llmint88bitmatrixmultiplication, dettmers2023qloraefficientfinetuningquantized}, degrade low-bit training stability \citep{Fishman_Chmiel_Banner_Soudry_2024_ScalingFP8}, and motivate increasingly complex mitigation strategies such as double quantization, activation smoothing, Hadamard transforms, or fine-grained quantization schemes \citep{xiao2024smoothquantaccurateefficientposttraining, wang2025bitnetv2native4bit, deepseekai2025deepseekv3technicalreport}. By eliminating attention sinks and suppressing extreme activations, softpick directly addresses the root cause of these issues. Hence, models using softpick retain accuracy under aggressive quantization and benefit at lower bit widths, suggesting that simpler quantization and low-precision training pipelines may become viable without specialized outlier-handling mechanisms.

The sparsity induced in attention maps could enable potential inference acceleration via sparse kernels and reduce unnecessary computation in attention-value products. This sparsity also clarifies head behavior: heads that remain inactive until triggered are easier to identify and safely prune, potentially extending prior work on active-dormant heads \citep{guo2024activedormant} and head pruning \citep{Wang_2021, Shim_2021}. Finally, the resulting attention maps are more interpretable and visually legible, which may benefit analysis techniques such as attention rollout \citep{abnar-zuidema-2020-quantifying}. Since attention sinks and activation outliers are not unique to language models, softpick can also generalize to vision, video, and multimodal transformers, where similar artifacts have been observed \citep{darcet2024visiontransformersneedregisters, wen2025analysis, Kumar2024XLSRTransducerSA}.

\section{Conclusion}
We introduced Softpick, a rectified, non-sum-to-one drop-in replacement for softmax in transformer attention. Across 340M and 1.8B models trained from scratch, softpick eliminates attention sinks (0\% sink rate), induces genuinely sparse attention maps, and strongly suppresses massive activation outliers, leading to improved robustness under low-bit post-training quantization, especially at very low precisions. While results at 340M are competitive with softmax, scaling and long-context behavior remain open challenges, motivating future work on softmax alternatives with desirable properties for transformer language models.

\section*{Limitations}
First, softpick does not yet scale reliably under our current training recipe: while it is competitive with softmax at 340M, it underperforms at 1.8B when we reuse the same setup. We hypothesize that either a) underscoring of long contexts becomes severe at larger scales, resulting in poor context retrieval or b) during training, some attention heads may become persistently inactive (dead), reducing the model's effective capacity and contributing to the performance gap at larger scale. Additionally, as we explain in \S\ref{sec:long_context}, softpick models do not yet yield improved long-context retrieval in our experiments (e.g., passkey retrieval). Despite producing sharper and sparser attention maps, the results suggest that additional work is needed to maintain strong value signals as context length increases. Lastly, given a fixed compute budget, we prioritized experimental breadth with training the full 340M/1.8B models, running additional short-run ablations across multiple sink-mitigation variants and softpick-related modifications, and acquiring metrics for analysis, over exhaustive large-scale hyperparameter sweeps and substantially longer training runs; we therefore leave a more comprehensive scaling study (larger parameter counts, longer contexts, and longer training horizons) to future work.





\section*{Acknowledgments}
We would like to thank fal.ai and their grant program (\url{https://fal.ai/grants}) for providing us with the GPU hours needed to train all 340M and 1.8B models.
\bibliography{custom}

\newpage
\appendix
\onecolumn

\section{FlashAttention Algorithms}
\label{sec:flashattention-appendix}
\newcommand{\absmath}[1]{\left\lvert#1\right\rvert}
\newcommand{\norm}[1]{\left\|{#1}\right\|} 
\newcommand{\diag}{\mathrm{diag}}
\newcommand{\softmaxmath}{\mathrm{softmax}}
\newcommand{\dsoftmax}{\mathrm{dsoftmax}}
\providecommand{\tr}{\mathop{\rm tr}}

\newcommand{\defeq}{:=}

\newcommand{\vQ}{\mathbf{Q}}
\newcommand{\vK}{\mathbf{K}}
\newcommand{\vV}{\mathbf{V}}
\newcommand{\vdQ}{\mathbf{dQ}}
\newcommand{\vdK}{\mathbf{dK}}
\newcommand{\vdV}{\mathbf{dV}}
\newcommand{\vS}{\mathbf{S}}
\newcommand{\vdS}{\mathbf{dS}}
\newcommand{\vP}{\mathbf{P}}
\newcommand{\vdP}{\mathbf{dP}}
\newcommand{\vR}{\mathbf{R}}
\newcommand{\vdR}{\mathbf{dR}}
\newcommand{\vA}{\mathbf{A}}
\newcommand{\vdA}{\mathbf{dA}}
\newcommand{\vE}{\mathbf{E}}
\newcommand{\vU}{\mathbf{U}}
\newcommand{\vW}{\mathbf{W}}
\newcommand{\vT}{\mathbf{T}}
\newcommand{\vX}{\mathbf{X}}
\newcommand{\vO}{\mathbf{O}}
\newcommand{\vdO}{\mathbf{dO}}
\newcommand{\vM}{\mathbf{M}}
\newcommand{\vZ}{\mathbf{Z}}

\begin{algorithm}[H]
  \caption{\small\label{alg:flash2_fwd} FlashAttention-2 Forward Pass with Softpick}
  \begin{algorithmic}[1]
    \REQUIRE Matrices $\vQ, \vK, \vV \in \mathbb{R}^{N \times d}$ in HBM, block sizes $B_c$, $B_r$, denominator epsilon $\epsilon$
    \STATE \label{alg:stream_attn_split_qkv} Divide $\vQ$ into $T_r = \left\lceil\frac{N}{B_r} \right\rceil$ blocks $\vQ_1, \dots, \vQ_{T_r}$ of size $B_r \times d$ each,
    and divide $\vK, \vV$ in to $T_c = \left\lceil \frac{N}{B_c} \right\rceil$ blocks $\vK_1, \dots, \vK_{T_c}$ and
    $\vV_1, \dots, \vV_{T_c}$, of size $B_c \times d$ each.
    \STATE Divide the output $\vO \in \mathbb{R}^{N \times d}$ into $T_r$ blocks $\vO_i, \dots, \vO_{T_r}$ of size
    $B_r \times d$ each, and divide the logsumexp $L$ into $T_r$ blocks $L_i, \dots, L_{T_r}$ of size
    $B_r$ each.
    \FOR{$1 \le i \le T_r$} \label{alg:stream_attn_outer_loop}
      \STATE \label{alg:stream_attn_load_q} Load $\vQ_i$ from HBM to on-chip SRAM.
      \STATE \label{alg:stream_attn_init} On chip, initialize $\vO_{i}^{(0)} = (0)_{B_r \times d} \in \mathbb{R}^{B_r \times d}, \ell_{i}^{(0)} = (0)_{B_r} \in \mathbb{R}^{B_r}, m_{i}^{(0)} = (-\infty)_{B_r} \in \mathbb{R}^{B_r}$.
      \FOR{$1 \le j \le T_c$}
        \STATE \label{alg:stream_attn_load_kv} Load $\vK_j, \vV_j$ from HBM to on-chip SRAM.
        \STATE \label{alg:stream_attn_qk} On chip, compute $\vS_{i}^{(j)} = \vQ_i \vK_j^T \in \mathbb{R}^{B_r \times B_c}$.
        \STATE \label{alg:stream_attn_statistics} On chip, compute
        $m_{i}^{(j)} = \mathrm{max}(m_{i}^{(j-1)}, \mathrm{rowmax}(\vS_{i}^{(j)})) \in \mathbb{R}^{B_r}$,\\
        $\tilde{\vP}_{i}^{(j)} = \exp(\vS_{i}^{(j)} - m_{i}^{(j)}) - \exp(-m_{i}^{(j)})  \in \mathbb{R}^{B_r \times B_c}$ (pointwise), \\
        $\tilde{\vR}_{i}^{(j)} = ReLU(\tilde{\vP}_{i}^{(j)}) \in \mathbb{R}^{B_r \times B_c}$ (pointwise), \\
        $\tilde{\vA}_{i}^{(j)} = |\tilde{\vP}_{i}^{(j)}| \in \mathbb{R}^{B_r \times B_c}$ (pointwise), \\
        $\ell_{i}^{(j)} = e^{m_{i}^{j-1} - m_{i}^{(j)}} \ell_{i}^{(j-1)} + \mathrm{row sum}(\tilde{\vA}_{i}^{(j)}) \in \mathbb{R}^{B_r}$.
        \STATE \label{alg:stream_attn_update} On chip, compute
        $\vO_{i}^{(j)} = \diag(e^{m_{i}^{(j-1)} - m_{i}^{(j)}})^{-1} \vO_{i}^{(j-1)} + \tilde{\vR}_{i}^{(j)} \vV_j$.
      \ENDFOR
      \STATE On chip, compute $\ell_{i}^{(T_c)} = \ell_{i}^{(T_c)}+\epsilon$.
      \STATE On chip, compute $\vO_{i} = \diag(\ell_{i}^{(T_c)})^{-1} \vO_{i}^{(T_c)}$.
      \STATE On chip, compute $L_{i} = m_{i}^{(T_c)} + \log(\ell_i^{(T_c)})$.
      \STATE Write $\vO_{i}$ to HBM as the $i$-th block of $\vO$.
      \STATE Write $L_{i}$ to HBM as the $i$-th block of $L$.
    \ENDFOR
    \STATE Return the output $\vO$ and the logsumexp $L$.
  \end{algorithmic}
\end{algorithm}

\newpage
\begin{algorithm}[h]
  \caption{\small\label{alg:flash_bwd} FlashAttention-2 Backward Pass with Softpick}
  \begin{algorithmic}[1]
    \REQUIRE Matrices $\vQ, \vK, \vV, \vO, \vdO \in \mathbb{R}^{N \times d}$ in HBM,
    vector $L \in \mathbb{R}^N$ in HBM, block sizes $B_c$, $B_r$.
    \STATE Divide $\vQ$ into $T_r = \left\lceil\frac{N}{B_r} \right\rceil$ blocks $\vQ_1, \dots, \vQ_{T_r}$ of size $B_r \times d$ each,
    and divide $\vK, \vV$ in to $T_c = \left\lceil \frac{N}{B_c} \right\rceil$ blocks $\vK_1, \dots, \vK_{T_c}$ and
    $\vV_1, \dots, \vV_{T_c}$, of size $B_c \times d$ each.
    \STATE Divide $\vO$ into $T_r$ blocks $\vO_i, \dots, \vO_{T_r}$ of size
    $B_r \times d$ each, divide $\vdO$ into $T_r$ blocks $\vdO_i, \dots, \vdO_{T_r}$
    of size $B_r \times d$ each, and divide $L$ into $T_r$ blocks $L_i, \dots, L_{T_r}$ of size
    $B_r$ each.
    \STATE Initialize $\vdQ = (0)_{N \times d}$ in HBM and divide it into $T_r$ blocks $\vdQ_1, \dots, \vdQ_{T_r}$ of size $B_r \times d$ each.
    Divide $\vdK, \vdV \in \mathbb{R}^{N \times d}$ in to $T_c$ blocks $\vdK_1, \dots, \vdK_{T_c}$ and
    $\vdV_1, \dots, \vdV_{T_c}$, of size $B_c \times d$ each.
    \STATE Compute $D = \mathrm{rowsum}(\vdO \circ \vO) \in \mathbb{R}^d$ (pointwise multiply), write
    $D$ to HBM and divide it into $T_r$ blocks $D_1, \dots, D_{T_r}$ of size
    $B_r$ each.
    \FOR{$1 \le j \le T_c$}
      \STATE Load $\vK_j, \vV_j$ from HBM to on-chip SRAM.
      \STATE Initialize $\vdK_j = (0)_{B_c \times d}, \vdV_j = (0)_{B_c \times d}$ on SRAM.
      \FOR{$1 \le i \le T_r$}
        \STATE Load $\vQ_i, \vO_i, \vdO_i, \vdQ_i, L_i, D_i$ from HBM to on-chip SRAM.
        \STATE On chip, compute $\vS_{i}^{(j)} = \vQ_i \vK_j^T \in \mathbb{R}^{B_r \times B_c}$.
        \STATE On chip, compute $\vE_{i}^{(j)} = \exp(\vS_{ij} - L_{i}) \in \mathbb{R}^{B_r \times B_c}$, \\
        $\vP_{i}^{(j)} = \vE_{i}^{(j)} - \exp(-L_{i})\in \mathbb{R}^{B_r \times B_c}$, \\
        $\vR_{i}^{(j)} = ReLU(\vP_{i}^{(j)}) \in \mathbb{R}^{B_r \times B_c}$. \\
        \STATE On chip, compute
        $\vdV_j \leftarrow \vdV_j + (\vR_{i}^{(j)})^\top \vdO_i \in \mathbb{R}^{B_c \times d}$.
        \STATE On chip, compute
        $\vdP_{i}^{(j)} = \vdO_{i} \vV_j^\top \in \mathbb{R}^{B_r \times B_c}$, \\
        $\vdR_{i}^{(j)} = step(\vS_{i}^{(j)}) \circ \vdP_{i}^{(j)} \in \mathbb{R}^{B_r \times B_c}$ (pointwise multiply or use where() function), \\
        $\vdA_{i}^{(j)} = sign(\vS_{i}^{(j)}) \circ D_i \in \mathbb{R}^{B_r \times B_c}$ (pointwise multiply or use where() function).
        \STATE On chip, compute $\vdS_{i}^{(j)} = \vE_{i}^{(j)} \circ (\vdR_{i}^{(j)} - \vdA_{i}^{(j)}) \in \mathbb{R}^{B_r \times B_c}$.
        \STATE Load $\vdQ_i$ from HBM to SRAM, then on chip, update
        $\vdQ_{i} \leftarrow \vdQ_i + \vdS_{i}^{(j)} \vK_j \in \mathbb{R}^{B_r \times d}$, and write
        back to HBM.
        \STATE On chip, compute $\vdK_{j} \leftarrow \vdK_j + {\vdS_{i}^{(j)}}^\top \vQ_i \in \mathbb{R}^{B_c \times d}$.
      \ENDFOR
      \STATE Write $\vdK_j, \vdV_j$ to HBM.
    \ENDFOR
    \STATE Return $\vdQ, \vdK, \vdV$.
  \end{algorithmic}
\end{algorithm}

\newpage
\clearpage
\twocolumn
\section{Training Configuration}
\label{sec:training-configuration}


\begin{table}[H]
  \centering
  \small
  \setlength{\tabcolsep}{3pt}
  \renewcommand{\arraystretch}{1.08}
  \begin{tabular}{@{}l l @{\hspace{4pt}} l l@{}}
    \toprule
    Param. & Value. & Param. & Value. \\
    \midrule

    \multicolumn{4}{@{}l}{\textbf{Architecture}} \\
    \addlinespace[2pt]
    Hidden size   & \{1024, 2048\} & \# Heads    & \{16, 32\} \\
    \# Layers   & \{24, 32\}       & KV heads & \{16, 32\} \\
    Seq. length   & 4096           & RoPE $\theta$ & 10k \\
    Vocab size    & 32k            & Tied emb.     & False \\
    \addlinespace[4pt]

    \multicolumn{4}{@{}l}{\textbf{Optimization \& training}} \\
    \addlinespace[2pt]
    Optimizer     & AdamW          & LR schedule   & cosine \\
    LR            & \{3e-4, 2e-4\}  & Min LR (\%)   & 10\% \\
    Warmup        & \{1k, 2k\}      & Train steps   & \{100k, 200k\} \\
    Dev. batch    & \{16, 8\}       & Analysis steps &  10k \\
    Grad. accum.  & \{1, 2\}        &  Global batch& 128\\
    Softpick $\epsilon$ & 1e-6      &  Max grad norm            &  1.0     \\
    \bottomrule
  \end{tabular}
  \caption{Training configuration and hyperparameters for \{340M, 1.8B\} models.}
  \label{tab:training-config}
\end{table}

\section{Other Experiments}
\label{sec:other-experiments-appendix}
\newcommand{\Attn}{\mathrm{Attn}}
\newcommand{\SoftpickOp}{\mathrm{Softpick}}
We report negative results from two additional variants related to Softpick. 
First, we replace the standard softmax with a \emph{rectified-only softmax}, which masks negative logits before normalization:
\begin{align*}
\nu(x) &=
\begin{cases}
x & x \ge 0,\\
-\infty & x < 0,
\end{cases}\\
\softmax(\mathbf{x})_i &= \frac{e^{\nu(x_i)}}{\sum_{j=1}^{N} e^{\nu(x_j)}}.
\end{align*}

Second, we evaluate a \emph{scalable-softpick} variant inspired by \citet{nakanishi2025scalablesoftmax}, which introduces a sequence-length-dependent scaling factor:
\begin{equation}
\alpha \;=\; \frac{s\log n}{\sqrt{d_k}},
\label{eq:softpick-alpha}
\end{equation}
where $s$ is a per-head trainable parameter and $n$ is the sequence length. In our implementation, $n$ is a vector that assigns a (possibly different) effective sequence length to each token position. The resulting attention becomes
\begin{equation}
\Attn(\mathbf{Q},\mathbf{K},\mathbf{V})
= \SoftpickOp(\alpha\,\mathbf{Q}\mathbf{K}^T)\,\mathbf{V}.
\label{eq:scalable-softpick-attn}
\end{equation}

\begin{figure}
    \centering
    \includegraphics[width=0.95\textwidth]{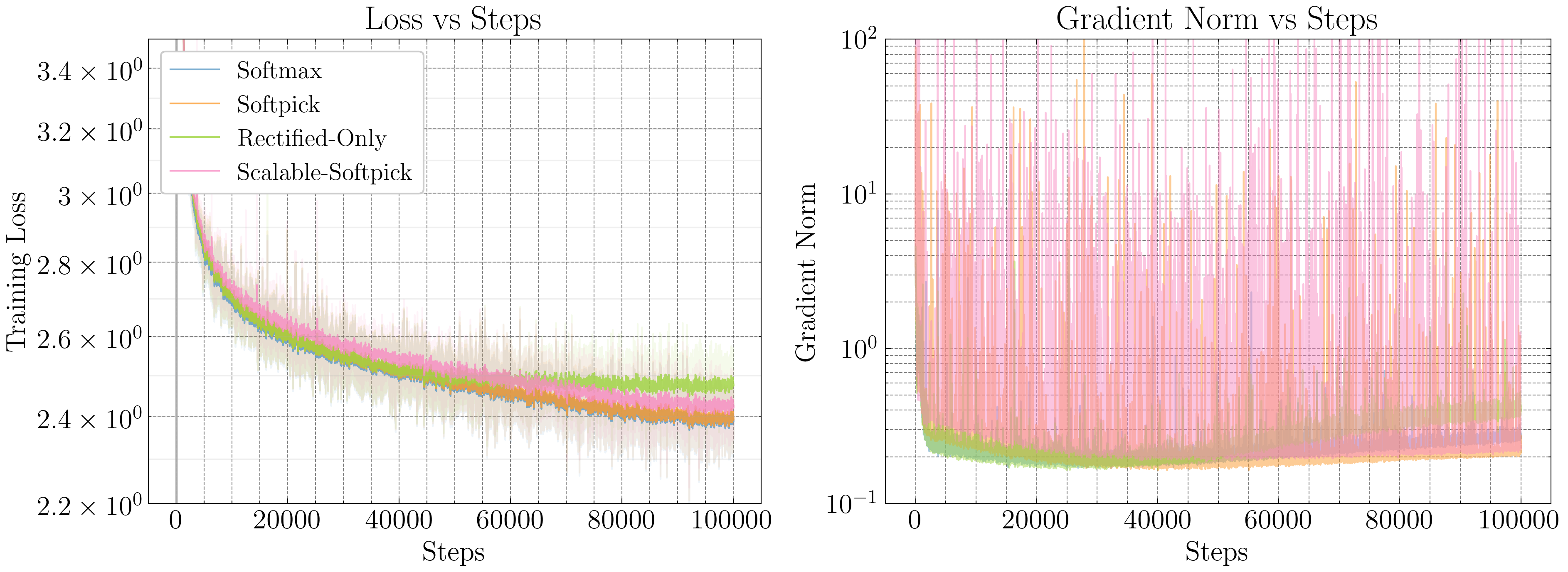}
    \caption{Training loss and gradient norm during training, including other experiments.}
    \label{fig:other-loss}
\end{figure}

We train both variants using the same setup and configuration as our main 340M-parameter models. 
We report training loss and gradient-norm curves in Figure~\ref{fig:other-loss}, benchmark results in Table~\ref{tab:other-benchmark-table}, and passkey retrieval results in Table~\ref{tab:other-passkey-table}.

\clearpage
\onecolumn
\begin{table} 
\centering
\begin{tabular}{l|p{3.0cm}|cccc}
\toprule
Task & Metric & Softmax & Softpick & Rectified-Only & Scalable-Softpick  \\
\midrule
\multirow[c]{2}{*}{Arc Easy} & Acc Norm \hfill $\uparrow$ & 56.31 & 56.61 & 51.52 & 55.89 \\
 & Acc \hfill $\uparrow$ & 60.61 & 60.82 & 57.66 & 60.98 \\
\cmidrule{1-6}
\multirow[c]{2}{*}{Lambada} & Acc \hfill $\uparrow$ & 36.35 & 36.21 & 29.05 & 33.17 \\
 & Perplexity \hfill $\downarrow$ & 30.33 & 28.67 & 53.77 & 34.49 \\
\cmidrule{1-6}
\multirow[c]{2}{*}{Piqa} & Acc Norm \hfill $\uparrow$ & 66.43 & 66.27 & 66.27 & 66.32 \\
 & Acc \hfill $\uparrow$ & 66.87 & 66.43 & 66.43 & 66.92 \\
\cmidrule{1-6}
\multirow[c]{2}{*}{Sciq} & Acc Norm \hfill $\uparrow$ & 75.10 & 77.20 & 68.80 & 75.00 \\
 & Acc \hfill $\uparrow$ & 83.30 & 83.60 & 78.80 & 83.70 \\
\cmidrule{1-6}
Wikitext & Word Perplexity \hfill $\downarrow$ & 24.01 & 24.41 & 27.93 & 25.13 \\
\bottomrule
\end{tabular}
\caption{Comparison of softmax, softpick, rectified-only softmax, and scalable-softpick performance on downstream tasks. $\uparrow$ = Higher is better, $\downarrow$ = Lower is better.}
\vspace{5pt}
\label{tab:other-benchmark-table}

\end{table}

\clearpage
\newpage
\section{Full Results on Quantization of the 340M Models}
\label{sec:quantization-appendix}

\begin{table}[htbp!] 
\centering
\caption{Comparison of softpick vs softmax performance for BNB \cite{dettmers2022llmint88bitmatrixmultiplication,dettmers2023qloraefficientfinetuningquantized} quantization methods. $\uparrow$=Higher is better, $\downarrow$=Lower is better. $\Delta$ = Softpick - Softmax.}
\label{tab:softpick_vanilla_quant_filtered_bnb_340M}
\begin{tabular}{l|l|c|rrr}
\toprule
Task & Metric & Quantization & Softmax & Softpick & $\Delta$ \\
\midrule
\multirow[c]{4}{*}{Arc Easy} & Acc Norm \hfill $\uparrow$ & 4-bit & 53.66 & 53.87 & \textcolor{green!60!black}{+0.21} \\
\cmidrule{2-6}
 & Acc \hfill $\uparrow$ & 4-bit & 58.88 & 59.60 & \textcolor{green!60!black}{+0.72} \\
\cmidrule{2-6}
 & Acc Norm \hfill $\uparrow$ & 8-bit & 55.51 & 56.36 & \textcolor{green!60!black}{+0.84} \\
\cmidrule{2-6}
 & Acc \hfill $\uparrow$ & 8-bit & 59.81 & 60.82 & \textcolor{green!60!black}{+1.01} \\
\cmidrule{1-6}
\multirow[c]{4}{*}{Lambada} & Acc \hfill $\uparrow$ & 4-bit & 33.63 & 31.73 & \textcolor{red!70!black}{-1.90} \\
\cmidrule{2-6}
 & Perplexity \hfill $\downarrow$ & 4-bit & 37.59 & 39.25 & \textcolor{red!70!black}{+1.67} \\
\cmidrule{2-6}
 & Acc \hfill $\uparrow$ & 8-bit & 35.53 & 35.86 & \textcolor{green!60!black}{+0.33} \\
\cmidrule{2-6}
 & Perplexity \hfill $\downarrow$ & 8-bit & 31.22 & 28.74 & \textcolor{green!60!black}{-2.48} \\
\cmidrule{1-6}
\multirow[c]{4}{*}{Piqa} & Acc Norm \hfill $\uparrow$ & 4-bit & 65.29 & 66.38 & \textcolor{green!60!black}{+1.09} \\
\cmidrule{2-6}
 & Acc \hfill $\uparrow$ & 4-bit & 65.40 & 66.32 & \textcolor{green!60!black}{+0.92} \\
\cmidrule{2-6}
 & Acc Norm \hfill $\uparrow$ & 8-bit & 66.32 & 66.32 & \textcolor{gray}{+0.00} \\
\cmidrule{2-6}
 & Acc \hfill $\uparrow$ & 8-bit & 67.36 & 66.38 & \textcolor{red!70!black}{-0.98} \\
\cmidrule{1-6}
\multirow[c]{4}{*}{Sciq} & Acc Norm \hfill $\uparrow$ & 4-bit & 74.90 & 75.10 & \textcolor{green!60!black}{+0.20} \\
\cmidrule{2-6}
 & Acc \hfill $\uparrow$ & 4-bit & 82.90 & 82.80 & \textcolor{red!70!black}{-0.10} \\
\cmidrule{2-6}
 & Acc Norm \hfill $\uparrow$ & 8-bit & 75.20 & 77.40 & \textcolor{green!60!black}{+2.20} \\
\cmidrule{2-6}
 & Acc \hfill $\uparrow$ & 8-bit & 83.40 & 83.50 & \textcolor{green!60!black}{+0.10} \\
\cmidrule{1-6} 
\multirow[c]{2}{*}{Wikitext} & \multirow[c]{2}{*}{Word Perplexity \hfill $\downarrow$} & 4-bit & 26.38 & 26.47 & \textcolor{red!70!black}{+0.09} \\
\cmidrule{3-6}
 &  & 8-bit & 23.95 & 24.40 & \textcolor{red!70!black}{+0.46} \\
\cmidrule{1-6}
\end{tabular}

\end{table}
\newpage

\begin{table}[htbp!] 
\centering
\small
\caption{Comparison of softpick vs softmax performance for GPTQ \cite{frantar2023gptqaccurateposttrainingquantization} quantization. $\uparrow$=Higher is better, $\downarrow$=Lower is better. $\Delta$ = Softpick - Softmax.}
\label{tab:softpick_vanilla_quant_filtered_gptq_340M}
\begin{tabular}{l|l|c|rrr}
\toprule
Task & Metric & Quantization & Softmax & Softpick & $\Delta$  \\
\midrule
\multirow[c]{8}{*}{Arc Easy} & Acc Norm \hfill $\uparrow$ & 2-bit & 27.27 & 29.76 & \textcolor{green!60!black}{+2.48} \\
\cmidrule{2-6}
 & Acc \hfill $\uparrow$ & 2-bit & 27.02 & 29.88 & \textcolor{green!60!black}{+2.86} \\
\cmidrule{2-6}
 & Acc Norm \hfill $\uparrow$ & 3-bit & 48.57 & 51.81 & \textcolor{green!60!black}{+3.24} \\
\cmidrule{2-6}
 & Acc \hfill $\uparrow$ & 3-bit & 53.83 & 56.65 & \textcolor{green!60!black}{+2.82} \\
\cmidrule{2-6}
 & Acc Norm \hfill $\uparrow$ & 4-bit & 53.96 & 54.29 & \textcolor{green!60!black}{+0.34} \\
\cmidrule{2-6}
 & Acc \hfill $\uparrow$ & 4-bit & 59.93 & 59.26 & \textcolor{red!70!black}{-0.67} \\
\cmidrule{2-6}
 & Acc Norm \hfill $\uparrow$ & 8-bit & 56.31 & 56.36 & \textcolor{green!60!black}{+0.04} \\
\cmidrule{2-6}
 & Acc \hfill $\uparrow$ & 8-bit & 60.31 & 60.94 & \textcolor{green!60!black}{+0.63} \\
\cmidrule{1-6} 
\multirow[c]{8}{*}{Lambada} & Acc \hfill $\uparrow$ & 2-bit & 0.04 & 0.43 & \textcolor{green!60!black}{+0.39} \\
\cmidrule{2-6}
 & Perplexity \hfill $\downarrow$ & 2-bit & 930980.10 & 133928.18 & \textcolor{green!60!black}{-797051.92} \\
\cmidrule{2-6}
 & Acc \hfill $\uparrow$ & 3-bit & 23.71 & 29.07 & \textcolor{green!60!black}{+5.36} \\
\cmidrule{2-6}
 & Perplexity \hfill $\downarrow$ & 3-bit & 82.12 & 52.32 & \textcolor{green!60!black}{-29.80} \\
\cmidrule{2-6}
 & Acc \hfill $\uparrow$ & 4-bit & 33.55 & 34.97 & \textcolor{green!60!black}{+1.42} \\
\cmidrule{2-6}
 & Perplexity \hfill $\downarrow$ & 4-bit & 37.30 & 31.08 & \textcolor{green!60!black}{-6.21} \\
\cmidrule{2-6}
 & Acc \hfill $\uparrow$ & 8-bit & 36.44 & 36.15 & \textcolor{red!70!black}{-0.29} \\
\cmidrule{2-6}
 & Perplexity \hfill $\downarrow$ & 8-bit & 30.06 & 28.66 & \textcolor{green!60!black}{-1.41} \\
\cmidrule{1-6}
\multirow[c]{8}{*}{Piqa} & Acc Norm \hfill $\uparrow$ & 2-bit & 48.42 & 53.59 & \textcolor{green!60!black}{+5.17} \\
\cmidrule{2-6}
 & Acc \hfill $\uparrow$ & 2-bit & 52.12 & 54.52 & \textcolor{green!60!black}{+2.39} \\
\cmidrule{2-6}
 & Acc Norm \hfill $\uparrow$ & 3-bit & 63.76 & 64.64 & \textcolor{green!60!black}{+0.87} \\
\cmidrule{2-6}
 & Acc \hfill $\uparrow$ & 3-bit & 64.53 & 65.56 & \textcolor{green!60!black}{+1.03} \\
\cmidrule{2-6}
 & Acc Norm \hfill $\uparrow$ & 4-bit & 66.10 & 66.43 & \textcolor{green!60!black}{+0.33} \\
\cmidrule{2-6}
 & Acc \hfill $\uparrow$ & 4-bit & 65.89 & 66.59 & \textcolor{green!60!black}{+0.71} \\
\cmidrule{2-6}
 & Acc Norm \hfill $\uparrow$ & 8-bit & 66.59 & 66.54 & \textcolor{red!70!black}{-0.05} \\
\cmidrule{2-6}
 & Acc \hfill $\uparrow$ & 8-bit & 66.81 & 66.21 & \textcolor{red!70!black}{-0.60} \\
\cmidrule{1-6}
\multirow[c]{8}{*}{Sciq} & Acc Norm \hfill $\uparrow$ & 2-bit & 29.80 & 35.80 & \textcolor{green!60!black}{+6.00} \\
\cmidrule{2-6}
 & Acc \hfill $\uparrow$ & 2-bit & 27.80 & 34.80 & \textcolor{green!60!black}{+7.00} \\
\cmidrule{2-6}
 & Acc Norm \hfill $\uparrow$ & 3-bit & 73.50 & 75.90 & \textcolor{green!60!black}{+2.40} \\
\cmidrule{2-6}
 & Acc \hfill $\uparrow$ & 3-bit & 79.30 & 82.30 & \textcolor{green!60!black}{+3.00} \\
\cmidrule{2-6}
 & Acc Norm \hfill $\uparrow$ & 4-bit & 73.40 & 76.40 & \textcolor{green!60!black}{+3.00} \\
\cmidrule{2-6}
 & Acc \hfill $\uparrow$ & 4-bit & 81.00 & 82.70 & \textcolor{green!60!black}{+1.70} \\
\cmidrule{2-6}
 & Acc Norm \hfill $\uparrow$ & 8-bit & 75.00 & 77.60 & \textcolor{green!60!black}{+2.60} \\
\cmidrule{2-6}
 & Acc \hfill $\uparrow$ & 8-bit & 83.20 & 83.70 & \textcolor{green!60!black}{+0.50} \\
\cmidrule{1-6}
\multirow[c]{4}{*}{Wikitext} & \multirow[c]{4}{*}{Word Perplexity \hfill $\downarrow$} & 2-bit & 4081.76 & 893.35 & \textcolor{green!60!black}{-3188.41} \\
 \cmidrule{3-6}
 &  & 3-bit & 34.57 & 31.53 & \textcolor{green!60!black}{-3.04} \\
 \cmidrule{3-6}
 &  & 4-bit & 25.69 & 25.53 & \textcolor{green!60!black}{-0.16} \\
 \cmidrule{3-6}
 &  & 8-bit & 23.84 & 24.31 & \textcolor{red!70!black}{+0.47} \\
\cmidrule{1-6}
\end{tabular}

\end{table}

\newpage

\begin{table}[htbp!] 
\centering
\caption{Comparison of softpick vs softmax performance for HQQ \cite{HQQ_2023} quantization methods. $\uparrow$=Higher is better, $\downarrow$=Lower is better. $\Delta$ = Softpick - Softmax.}
\label{tab:softpick_vanilla_quant_filtered_hqq_1_8B}
\begin{tabular}{l|l|c|rrr}
\toprule
Task & Metric & Quantization & Softmax & Softpick & $\Delta$  \\
\midrule
\multirow[c]{4}{*}{Arc Easy} & Acc Norm \hfill $\uparrow$ & 2-bit & 26.68 & 36.36 & \textcolor{green!60!black}{+9.68} \\
\cmidrule{2-6}
 & Acc \hfill $\uparrow$ & 2-bit & 26.81 & 37.54 & \textcolor{green!60!black}{+10.73} \\
\cmidrule{2-6}
 & Acc Norm \hfill $\uparrow$ & 3-bit & 48.65 & 52.57 & \textcolor{green!60!black}{+3.91} \\
\cmidrule{2-6}
 & Acc \hfill $\uparrow$ & 3-bit & 52.78 & 57.95 & \textcolor{green!60!black}{+5.18} \\
\cmidrule{1-6} \cmidrule{2-6}
\multirow[c]{4}{*}{Lambada} & Acc \hfill $\uparrow$ & 2-bit & 0.00 & 5.76 & \textcolor{green!60!black}{+5.76} \\
\cmidrule{2-6}
 & Perplexity \hfill $\downarrow$ & 2-bit & 21857842.67 & 3741.81 & \textcolor{green!60!black}{-21854100.86} \\
\cmidrule{2-6}
 & Acc \hfill $\uparrow$ & 3-bit & 26.18 & 32.14 & \textcolor{green!60!black}{+5.96} \\
\cmidrule{2-6}
 & Perplexity \hfill $\downarrow$ & 3-bit & 76.81 & 41.85 & \textcolor{green!60!black}{-34.96} \\
\cmidrule{1-6}
\multirow[c]{4}{*}{Piqa} & Acc Norm \hfill $\uparrow$ & 2-bit & 50.38 & 55.11 & \textcolor{green!60!black}{+4.73} \\
\cmidrule{2-6}
 & Acc \hfill $\uparrow$ & 2-bit & 52.29 & 56.04 & \textcolor{green!60!black}{+3.75} \\
\cmidrule{2-6}
 & Acc Norm \hfill $\uparrow$ & 3-bit & 62.40 & 65.78 & \textcolor{green!60!black}{+3.37} \\
\cmidrule{2-6}
 & Acc \hfill $\uparrow$ & 3-bit & 63.55 & 66.00 & \textcolor{green!60!black}{+2.45} \\
\cmidrule{1-6} \cmidrule{2-6}
\multirow[c]{4}{*}{Sciq} & Acc Norm \hfill $\uparrow$ & 2-bit & 22.20 & 51.40 & \textcolor{green!60!black}{+29.20} \\
\cmidrule{2-6}
 & Acc \hfill $\uparrow$ & 2-bit & 21.60 & 53.60 & \textcolor{green!60!black}{+32.00} \\
\cmidrule{2-6}
 & Acc Norm \hfill $\uparrow$ & 3-bit & 71.60 & 72.10 & \textcolor{green!60!black}{+0.50} \\
\cmidrule{2-6}
 & Acc \hfill $\uparrow$ & 3-bit & 80.50 & 80.70 & \textcolor{green!60!black}{+0.20} \\
\cmidrule{1-6}
\multirow[c]{2}{*}{Wikitext} & \multirow[c]{2}{*}{Word Perplexity \hfill $\downarrow$} & 2-bit & 127870.20 & 2139.86 & \textcolor{green!60!black}{-125730.34} \\
\cmidrule{3-6}
 &  & 3-bit & 35.52 & 30.24 & \textcolor{green!60!black}{-5.29} \\
\bottomrule
\end{tabular}

\end{table}

\newpage
\section{Full Results on Quantization of the 1.8B Models}
\label{sec:quantization-appendix-1.8B}
\begin{table}[htbp!] 
\centering
\caption{Comparison of softpick vs softmax performance for BNB \cite{dettmers2022llmint88bitmatrixmultiplication,dettmers2023qloraefficientfinetuningquantized} quantization methods. $\uparrow$=Higher is better, $\downarrow$=Lower is better. $\Delta$ = Softpick - Softmax.}
\label{tab:softpick_vanilla_quant_filtered_bnb_1_8B}
\begin{tabular}{l|l|c|rrr}
\toprule
Task & Metric & Quantization & Softmax & Softpick & $\Delta$  \\
\midrule
\multirow[c]{4}{*}{Arc Easy} & Acc Norm \hfill $\uparrow$ & 4-bit & 66.75 & 58.84 & \textcolor{red!70!black}{-7.91} \\
\cmidrule{2-6}
 & Acc \hfill $\uparrow$ & 4-bit & 71.93 & 67.05 & \textcolor{red!70!black}{-4.88} \\
\cmidrule{2-6}
 & Acc Norm \hfill $\uparrow$ & 8-bit & 67.26 & 62.21 & \textcolor{red!70!black}{-5.05} \\
\cmidrule{2-6}
 & Acc \hfill $\uparrow$ & 8-bit & 72.22 & 69.07 & \textcolor{red!70!black}{-3.16} \\
\cmidrule{1-6} \cmidrule{2-6}
\multirow[c]{4}{*}{Lambada} & Acc \hfill $\uparrow$ & 4-bit & 47.84 & 41.26 & \textcolor{red!70!black}{-6.58} \\
\cmidrule{2-6}
 & Perplexity \hfill $\downarrow$ & 4-bit & 12.90 & 18.56 & \textcolor{red!70!black}{+5.66} \\
\cmidrule{2-6}
 & Acc \hfill $\uparrow$ & 8-bit & 48.65 & 43.14 & \textcolor{red!70!black}{-5.51} \\
\cmidrule{2-6}
 & Perplexity \hfill $\downarrow$ & 8-bit & 11.74 & 16.11 & \textcolor{red!70!black}{+4.36} \\
\cmidrule{1-6} \cmidrule{2-6}
\multirow[c]{4}{*}{Piqa} & Acc Norm \hfill $\uparrow$ & 4-bit & 72.14 & 70.57 & \textcolor{red!70!black}{-1.58} \\
\cmidrule{2-6}
 & Acc \hfill $\uparrow$ & 4-bit & 72.63 & 71.00 & \textcolor{red!70!black}{-1.63} \\
\cmidrule{2-6}
 & Acc Norm \hfill $\uparrow$ & 8-bit & 72.96 & 70.40 & \textcolor{red!70!black}{-2.56} \\
\cmidrule{2-6}
 & Acc \hfill $\uparrow$ & 8-bit & 73.56 & 71.22 & \textcolor{red!70!black}{-2.34} \\
\cmidrule{1-6} \cmidrule{2-6}
\multirow[c]{4}{*}{Sciq} & Acc Norm \hfill $\uparrow$ & 4-bit & 86.90 & 79.40 & \textcolor{red!70!black}{-7.50} \\
\cmidrule{2-6}
 & Acc \hfill $\uparrow$ & 4-bit & 90.90 & 86.90 & \textcolor{red!70!black}{-4.00} \\
\cmidrule{2-6}
 & Acc Norm \hfill $\uparrow$ & 8-bit & 86.00 & 79.90 & \textcolor{red!70!black}{-6.10} \\
\cmidrule{2-6}
 & Acc \hfill $\uparrow$ & 8-bit & 90.30 & 86.90 & \textcolor{red!70!black}{-3.40} \\
\cmidrule{1-6} \cmidrule{2-6}
\multirow[c]{2}{*}{Wikitext} & \multirow[c]{2}{*}{Word Perplexity \hfill $\downarrow$} & 4-bit & 16.17 & 18.97 & \textcolor{red!70!black}{+2.80} \\
 &  & 8-bit & 15.20 & 17.94 & \textcolor{red!70!black}{+2.74} \\
\bottomrule
\end{tabular}

\end{table}
\newpage
\begin{table}[htbp!] 
\centering
\caption{Comparison of softpick vs softmax performance for GPTQ \cite{frantar2023gptqaccurateposttrainingquantization} quantization. $\uparrow$=Higher is Better, $\downarrow$=Lower is Better. $\Delta$ = Softpick - Softmax.}
\label{tab:softpick_vanilla_quant_filtered_gptq_1_8B}
\begin{tabular}{l|l|c|rrr}
\toprule
Task & Metric & Quantization & Softmax & Softpick & $\Delta$  \\
\midrule
\multirow[c]{8}{*}{Arc Easy} & Acc Norm \hfill $\uparrow$ & 2-bit & 28.62 & 34.60 & \textcolor{green!60!black}{+5.98} \\
\cmidrule{2-6}
 & Acc \hfill $\uparrow$ & 2-bit & 29.67 & 33.88 & \textcolor{green!60!black}{+4.21} \\
\cmidrule{2-6}
 & Acc Norm \hfill $\uparrow$ & 3-bit & 61.28 & 57.49 & \textcolor{red!70!black}{-3.79} \\
\cmidrule{2-6}
 & Acc \hfill $\uparrow$ & 3-bit & 64.90 & 64.27 & \textcolor{red!70!black}{-0.63} \\
\cmidrule{2-6}
 & Acc Norm \hfill $\uparrow$ & 4-bit & 65.74 & 61.87 & \textcolor{red!70!black}{-3.87} \\
\cmidrule{2-6}
 & Acc \hfill $\uparrow$ & 4-bit & 70.20 & 68.73 & \textcolor{red!70!black}{-1.47} \\
\cmidrule{2-6}
 & Acc Norm \hfill $\uparrow$ & 8-bit & 66.96 & 61.95 & \textcolor{red!70!black}{-5.01} \\
\cmidrule{2-6}
 & Acc \hfill $\uparrow$ & 8-bit & 72.77 & 68.60 & \textcolor{red!70!black}{-4.17} \\
\cmidrule{1-6} \cmidrule{2-6}
\multirow[c]{8}{*}{Lambada} & Acc \hfill $\uparrow$ & 2-bit & 1.14 & 1.59 & \textcolor{green!60!black}{+0.45} \\
\cmidrule{2-6}
 & Perplexity \hfill $\downarrow$ & 2-bit & 27256.52 & 18730.23 & \textcolor{green!60!black}{-8526.29} \\
\cmidrule{2-6}
 & Acc \hfill $\uparrow$ & 3-bit & 40.71 & 36.77 & \textcolor{red!70!black}{-3.94} \\
\cmidrule{2-6}
 & Perplexity \hfill $\downarrow$ & 3-bit & 20.58 & 25.33 & \textcolor{red!70!black}{+4.75} \\
\cmidrule{2-6}
 & Acc \hfill $\uparrow$ & 4-bit & 49.64 & 40.95 & \textcolor{red!70!black}{-8.69} \\
\cmidrule{2-6}
 & Perplexity \hfill $\downarrow$ & 4-bit & 11.43 & 17.94 & \textcolor{red!70!black}{+6.50} \\
\cmidrule{2-6}
 & Acc \hfill $\uparrow$ & 8-bit & 49.60 & 43.41 & \textcolor{red!70!black}{-6.19} \\
\cmidrule{2-6}
 & Perplexity \hfill $\downarrow$ & 8-bit & 11.35 & 15.82 & \textcolor{red!70!black}{+4.47} \\
\cmidrule{1-6} \cmidrule{2-6}
\multirow[c]{8}{*}{Piqa} & Acc Norm \hfill $\uparrow$ & 2-bit & 53.97 & 53.92 & \textcolor{red!70!black}{-0.05} \\
\cmidrule{2-6}
 & Acc \hfill $\uparrow$ & 2-bit & 54.46 & 55.22 & \textcolor{green!60!black}{+0.76} \\
\cmidrule{2-6}
 & Acc Norm \hfill $\uparrow$ & 3-bit & 71.00 & 70.51 & \textcolor{red!70!black}{-0.49} \\
\cmidrule{2-6}
 & Acc \hfill $\uparrow$ & 3-bit & 69.91 & 70.18 & \textcolor{green!60!black}{+0.27} \\
\cmidrule{2-6}
 & Acc Norm \hfill $\uparrow$ & 4-bit & 73.34 & 69.97 & \textcolor{red!70!black}{-3.37} \\
\cmidrule{2-6}
 & Acc \hfill $\uparrow$ & 4-bit & 71.98 & 71.11 & \textcolor{red!70!black}{-0.87} \\
\cmidrule{2-6}
 & Acc Norm \hfill $\uparrow$ & 8-bit & 73.83 & 70.67 & \textcolor{red!70!black}{-3.16} \\
\cmidrule{2-6}
 & Acc \hfill $\uparrow$ & 8-bit & 73.72 & 71.27 & \textcolor{red!70!black}{-2.45} \\
\cmidrule{1-6} \cmidrule{2-6}
\multirow[c]{8}{*}{Sciq} & Acc Norm \hfill $\uparrow$ & 2-bit & 45.80 & 45.30 & \textcolor{red!70!black}{-0.50} \\
\cmidrule{2-6}
 & Acc \hfill $\uparrow$ & 2-bit & 43.90 & 43.80 & \textcolor{red!70!black}{-0.10} \\
\cmidrule{2-6}
 & Acc Norm \hfill $\uparrow$ & 3-bit & 82.20 & 78.30 & \textcolor{red!70!black}{-3.90} \\
\cmidrule{2-6}
 & Acc \hfill $\uparrow$ & 3-bit & 88.00 & 85.50 & \textcolor{red!70!black}{-2.50} \\
\cmidrule{2-6}
 & Acc Norm \hfill $\uparrow$ & 4-bit & 85.90 & 80.40 & \textcolor{red!70!black}{-5.50} \\
\cmidrule{2-6}
 & Acc \hfill $\uparrow$ & 4-bit & 90.50 & 86.60 & \textcolor{red!70!black}{-3.90} \\
\cmidrule{2-6}
 & Acc Norm \hfill $\uparrow$ & 8-bit & 86.40 & 80.40 & \textcolor{red!70!black}{-6.00} \\
\cmidrule{2-6}
 & Acc \hfill $\uparrow$ & 8-bit & 90.00 & 87.20 & \textcolor{red!70!black}{-2.80} \\
\cmidrule{1-6} \cmidrule{2-6}
\multirow[c]{4}{*}{Wikitext} & \multirow[c]{4}{*}{Word Perplexity \hfill $\downarrow$} & 2-bit & 4660.08 & 336.22 & \textcolor{green!60!black}{-4323.86} \\
 &  & 3-bit & 19.24 & 21.77 & \textcolor{red!70!black}{+2.52} \\
 &  & 4-bit & 15.84 & 18.46 & \textcolor{red!70!black}{+2.62} \\
 &  & 8-bit & 15.10 & 17.86 & \textcolor{red!70!black}{+2.77} \\
\bottomrule
\end{tabular}

\end{table}
\newpage
\begin{table}[htbp!] 
\centering
\caption{Comparison of softpick vs softmax performance for HQQ \cite{HQQ_2023} quantization methods. $\uparrow$=Higher is better, $\downarrow$=Lower is better. $\Delta$ = Softpick - Softmax.}
\label{tab:softpick_vanilla_quant_filtered_hqq_1_8B}
\begin{tabular}{l|l|c|rrr}
\toprule
Task & Metric & Quantization & Softmax & Softpick & $\Delta$  \\
\midrule
\multirow[c]{4}{*}{Arc Easy} & Acc Norm \hfill $\uparrow$ & 2-bit & 67.21 & 61.99 & \textcolor{red!70!black}{-5.22} \\
\cmidrule{2-6}
 & Acc \hfill $\uparrow$ & 2-bit & 72.64 & 68.60 & \textcolor{red!70!black}{-4.04} \\
\cmidrule{2-6}
 & Acc Norm \hfill $\uparrow$ & 3-bit & 67.21 & 61.99 & \textcolor{red!70!black}{-5.22} \\
\cmidrule{2-6}
 & Acc \hfill $\uparrow$ & 3-bit & 72.64 & 68.60 & \textcolor{red!70!black}{-4.04} \\
\cmidrule{1-6} \cmidrule{2-6}
\multirow[c]{4}{*}{Lambada} & Acc \hfill $\uparrow$ & 2-bit & 49.56 & 43.61 & \textcolor{red!70!black}{-5.96} \\
\cmidrule{2-6}
 & Perplexity \hfill $\downarrow$ & 2-bit & 11.38 & 15.83 & \textcolor{red!70!black}{+4.45} \\
\cmidrule{2-6}
 & Acc \hfill $\uparrow$ & 3-bit & 49.56 & 43.61 & \textcolor{red!70!black}{-5.96} \\
\cmidrule{2-6}
 & Perplexity \hfill $\downarrow$ & 3-bit & 11.38 & 15.83 & \textcolor{red!70!black}{+4.45} \\
\cmidrule{1-6} \cmidrule{2-6}
\multirow[c]{4}{*}{Piqa} & Acc Norm \hfill $\uparrow$ & 2-bit & 73.45 & 70.89 & \textcolor{red!70!black}{-2.56} \\
\cmidrule{2-6}
 & Acc \hfill $\uparrow$ & 2-bit & 73.78 & 71.38 & \textcolor{red!70!black}{-2.39} \\
\cmidrule{2-6}
 & Acc Norm \hfill $\uparrow$ & 3-bit & 73.45 & 70.89 & \textcolor{red!70!black}{-2.56} \\
\cmidrule{2-6}
 & Acc \hfill $\uparrow$ & 3-bit & 73.78 & 71.38 & \textcolor{red!70!black}{-2.39} \\
\cmidrule{1-6} \cmidrule{2-6}
\multirow[c]{4}{*}{Sciq} & Acc Norm \hfill $\uparrow$ & 2-bit & 86.30 & 80.40 & \textcolor{red!70!black}{-5.90} \\
\cmidrule{2-6}
 & Acc \hfill $\uparrow$ & 2-bit & 90.20 & 87.10 & \textcolor{red!70!black}{-3.10} \\
\cmidrule{2-6}
 & Acc Norm \hfill $\uparrow$ & 3-bit & 86.30 & 80.40 & \textcolor{red!70!black}{-5.90} \\
\cmidrule{2-6}
 & Acc \hfill $\uparrow$ & 3-bit & 90.20 & 87.10 & \textcolor{red!70!black}{-3.10} \\
\cmidrule{1-6} \cmidrule{2-6}
\multirow[c]{2}{*}{Wikitext} & \multirow[c]{2}{*}{Word Perplexity \hfill $\downarrow$} & 2-bit & 15.10 & 17.86 & \textcolor{red!70!black}{+2.77} \\
 &  & 3-bit & 15.10 & 17.86 & \textcolor{red!70!black}{+2.77} \\
\bottomrule
\end{tabular}

\end{table}

\clearpage
\twocolumn
\section{More Attention Maps}
\label{sec:attention-maps-appendix}
To show the difference between attention maps produced by softmax and softpick, we run 3 sample text inputs through the models and extract the resulting attention maps. The sample text inputs are the following:
\paragraph{Input 1:} "The dominant sequence transduction models are based on complex recurrent or convolutional neural networks that include an encoder and a decoder. The best performing models also connect the encoder and decoder through an attention mechanism."
\paragraph{Input 2:} "According to all known laws of aviation, there is no way that a bee should be able to fly. Its wings are too small to get its fat little body off the ground. The bee, of course, flies anyway. Because bees don't care what humans think is impossible."
\paragraph{Input 3:} "An idol whose dream is to become the owner of a fast food chain. Kiara is a phoenix, not a chicken or turkey (Very important). She burns brightly, working herself to the bone since she'll just be reborn from her ashes anyway."

\clearpage
\onecolumn
\begin{figure}
    \centering
    \includegraphics[width=\linewidth]{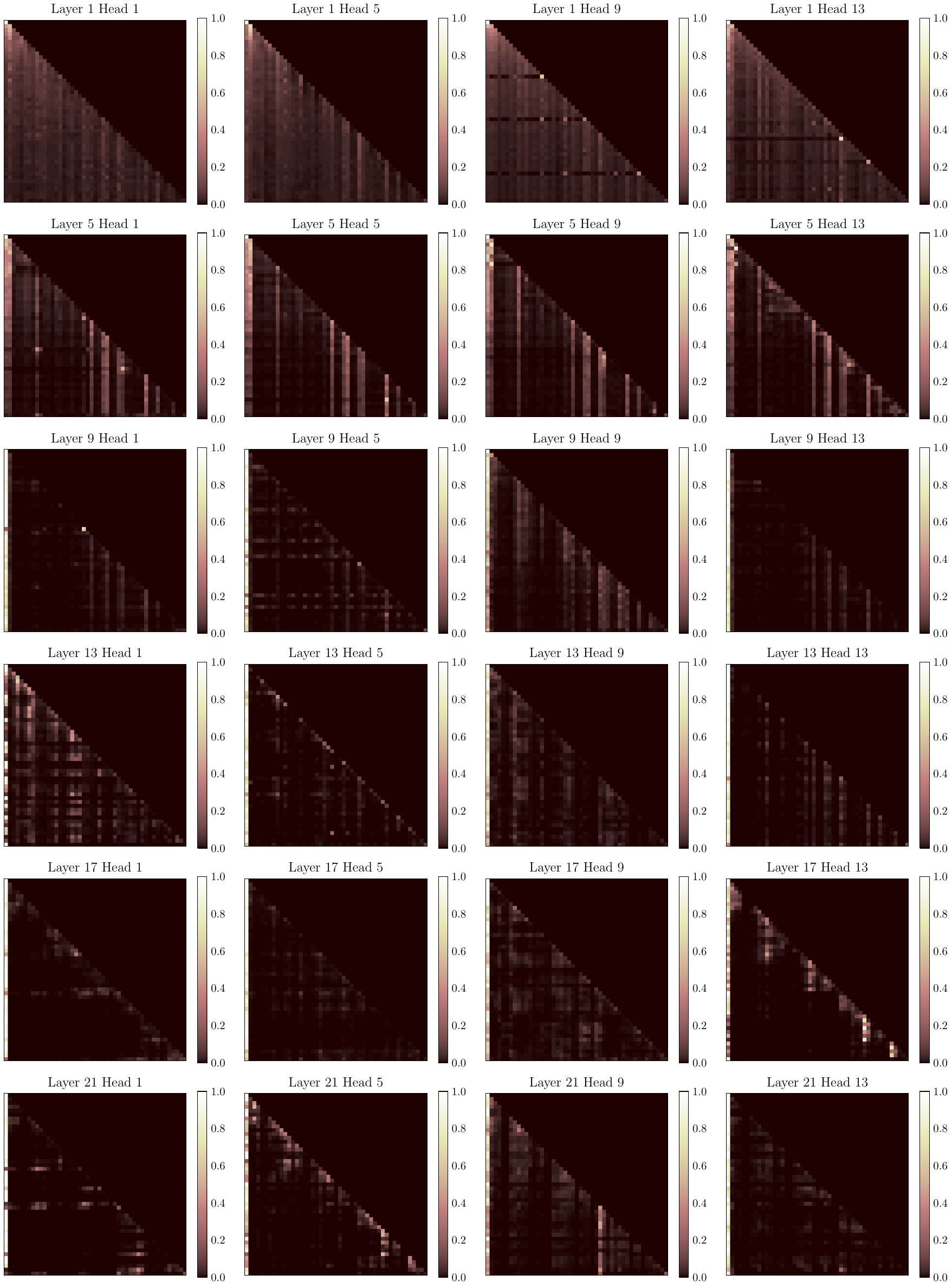}
    \caption{More attention maps of the softmax 340M model on sample text input 1.}
    \label{fig:input_1_softmax_map}
\end{figure}

\newpage
\begin{figure}
    \centering
    \includegraphics[width=\linewidth]{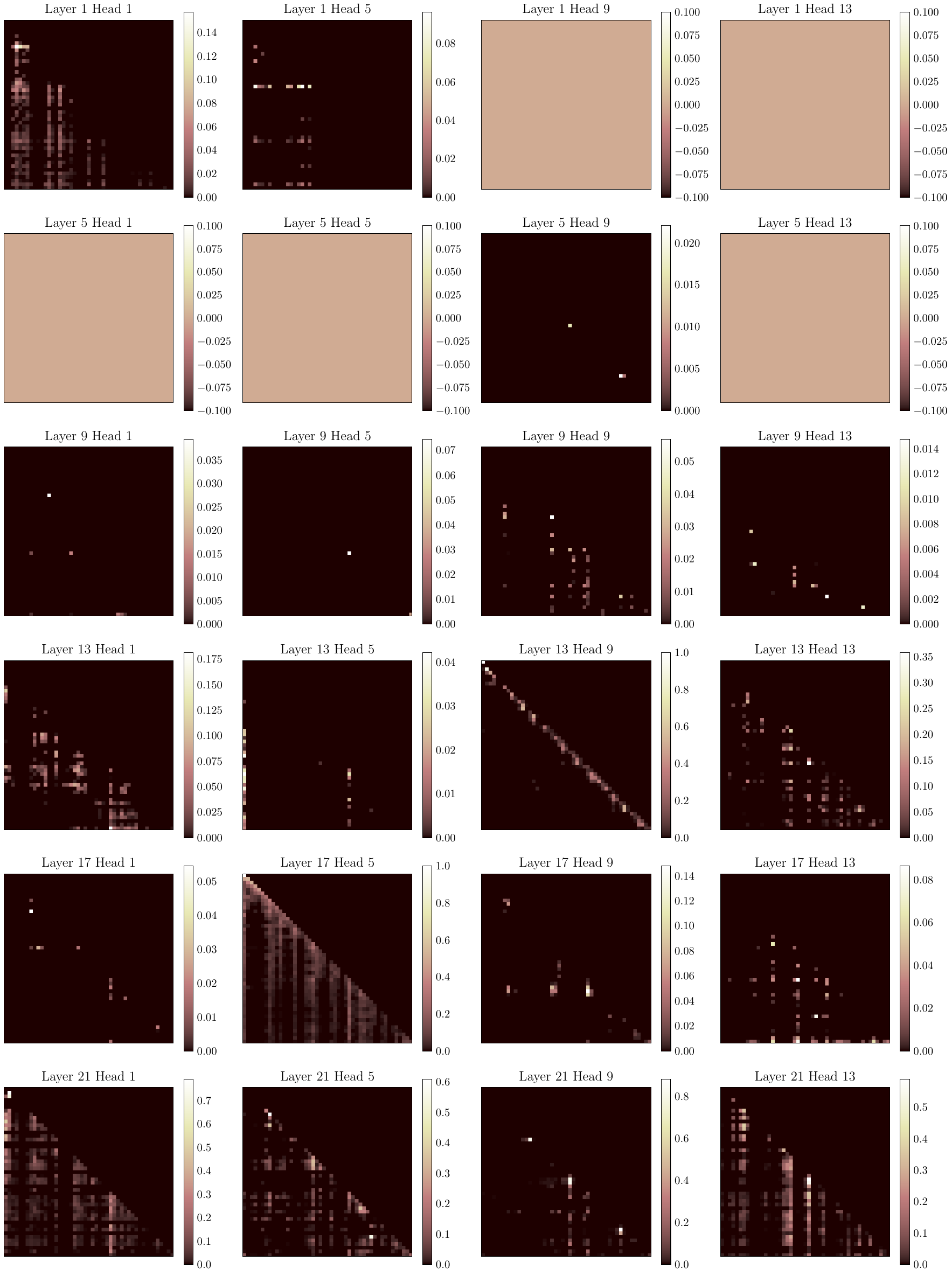}
    \caption{More attention maps of the softpick 340M model on sample text input 1.}
    \label{fig:input_1_softpick_map}
\end{figure}

\newpage
\begin{figure}
    \centering
    \includegraphics[width=\linewidth]{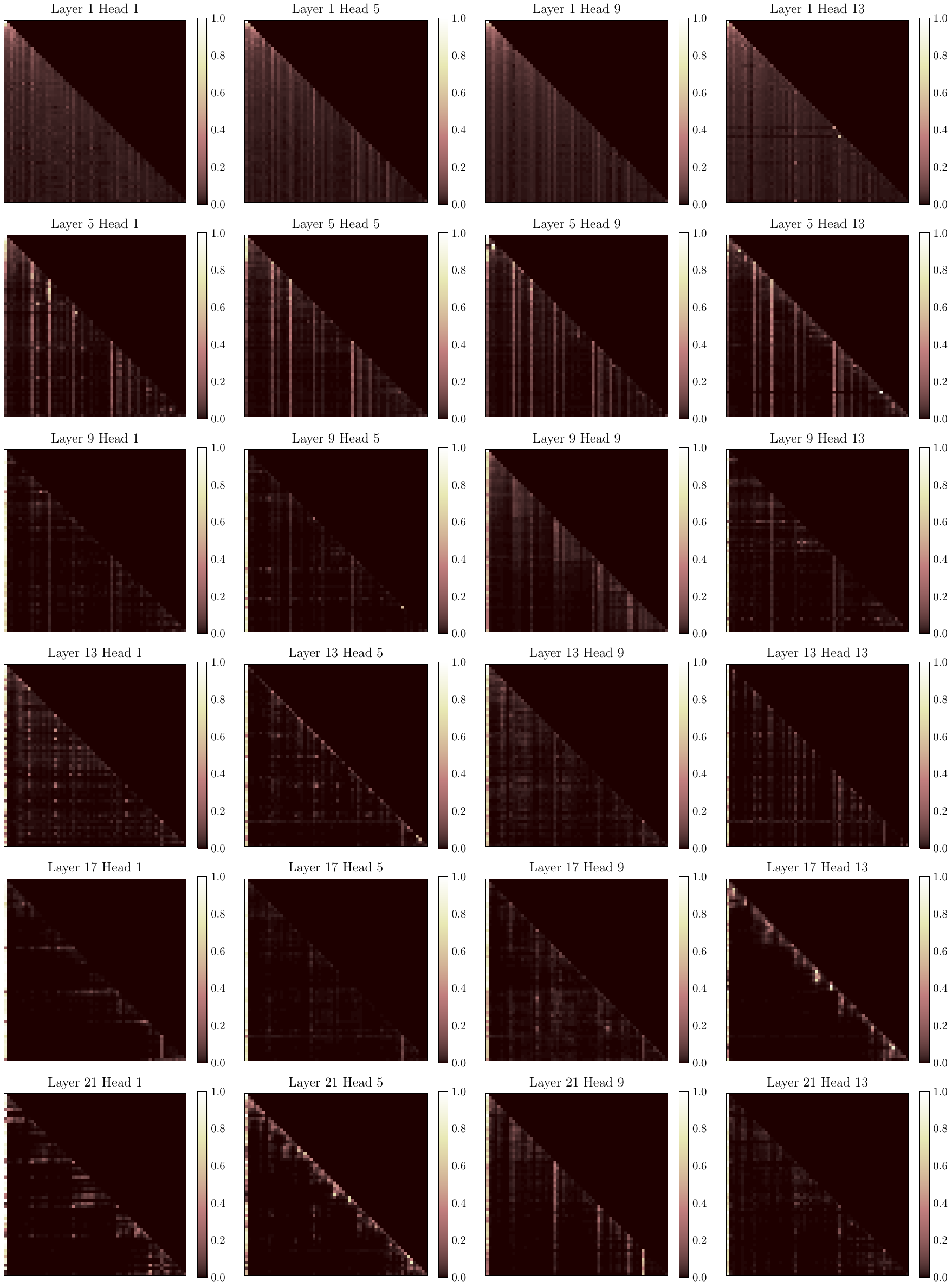}
    \caption{More attention maps of the softmax 340M model on sample text input 2.}
    \label{fig:input_2_softmax_map}
\end{figure}

\newpage
\begin{figure}
    \centering
    \includegraphics[width=\linewidth]{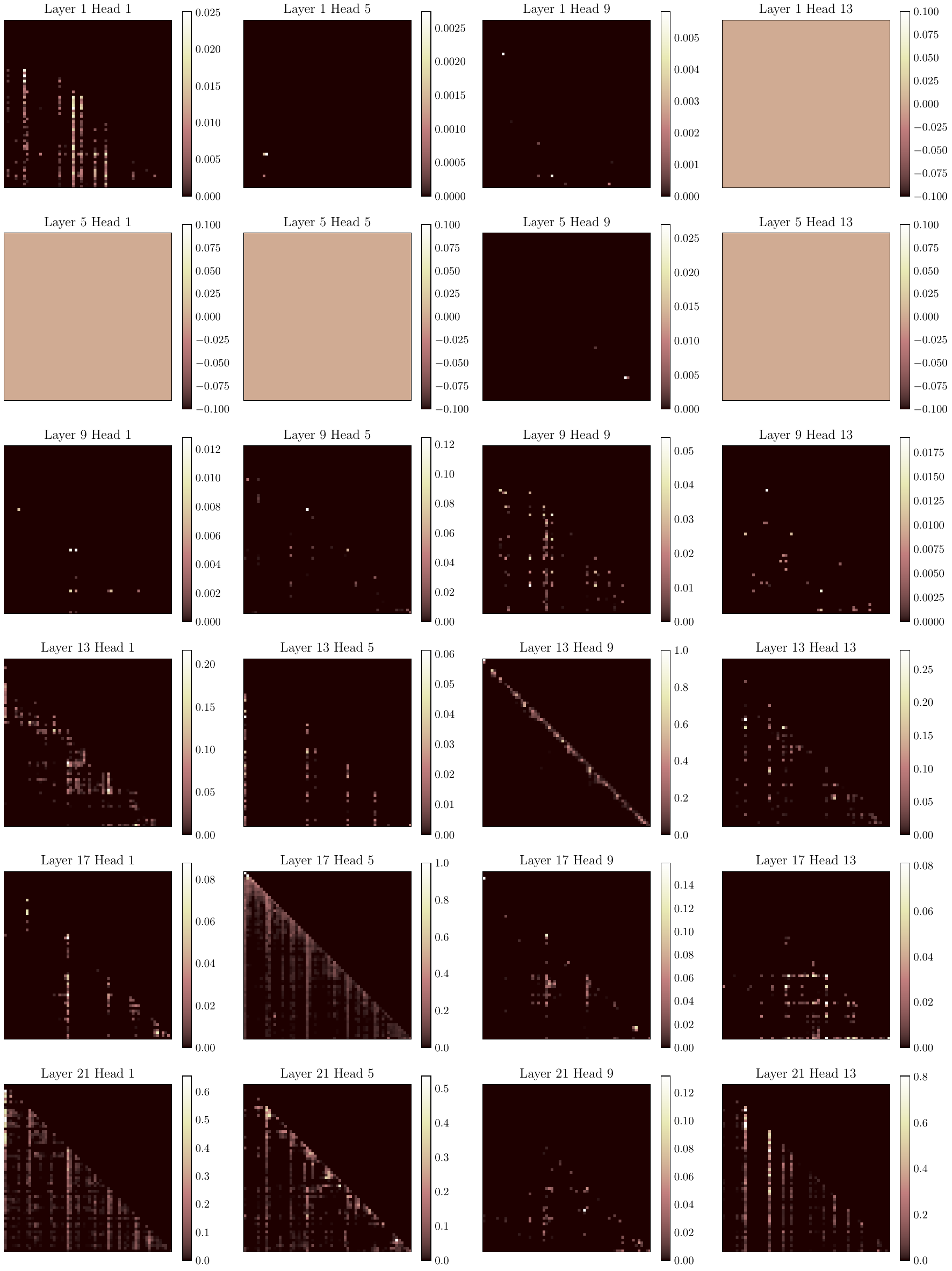}
    \caption{More attention maps of the softpick 340M model on sample text input 2.}
    \label{fig:input_2_softpick_map}
\end{figure}

\newpage
\begin{figure}
    \centering
    \includegraphics[width=\linewidth]{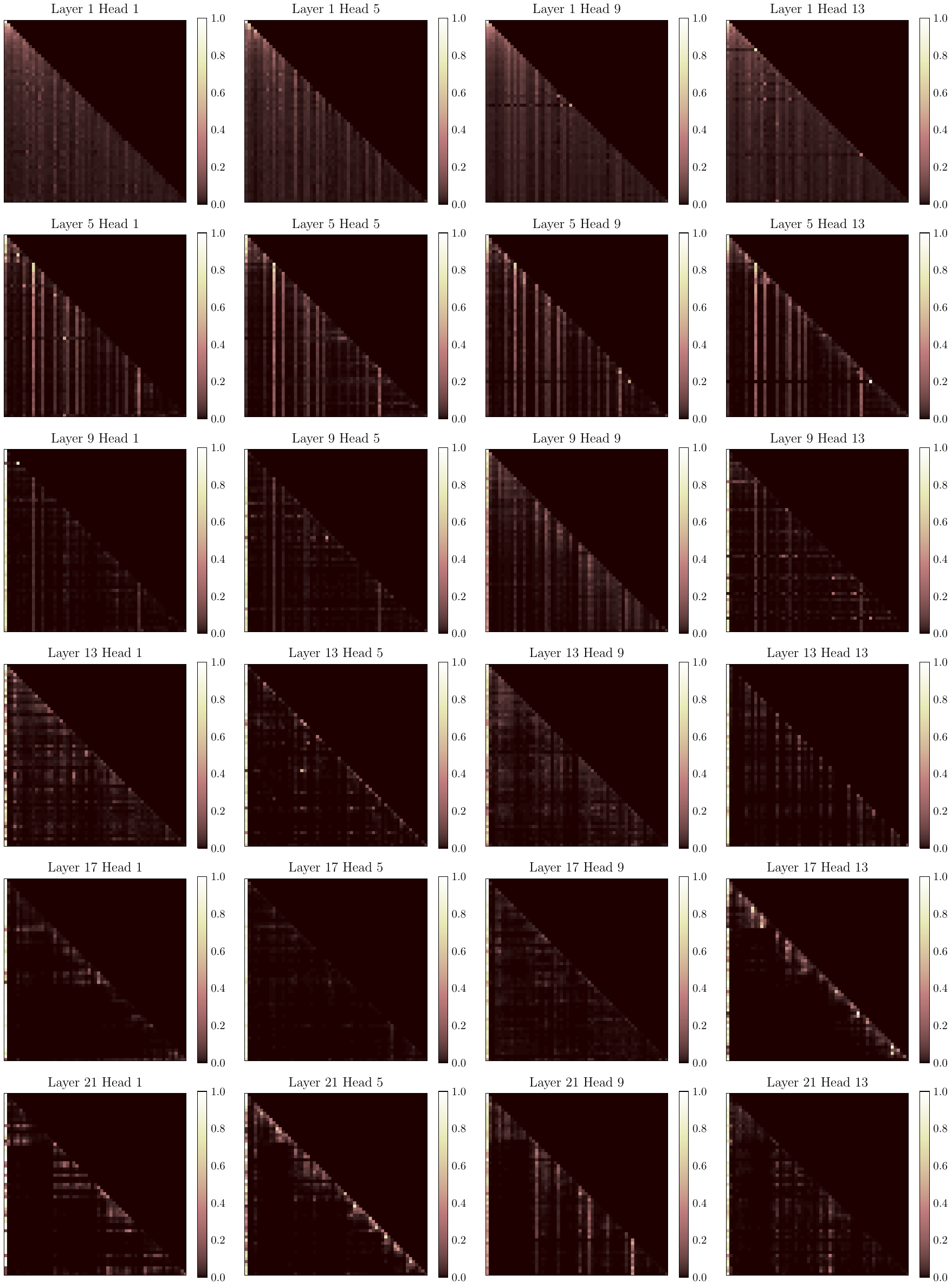}
    \caption{More attention maps of the softmax 340M model on sample text input 3.}
    \label{fig:input_3_softmax_map}
\end{figure}

\newpage
\begin{figure}
    \centering
    \includegraphics[width=\linewidth]{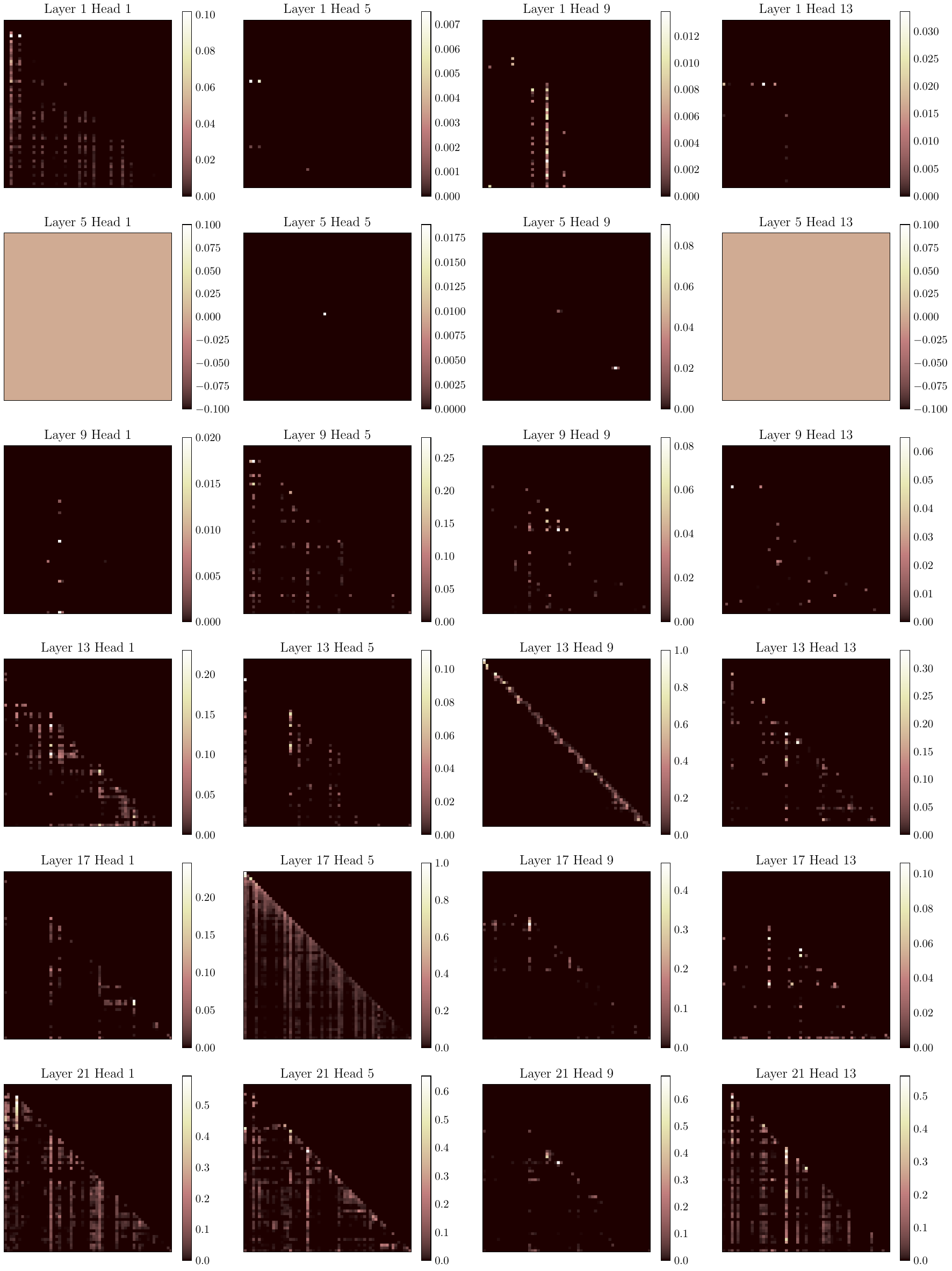}
    \caption{More attention maps of the softpick 340M model on sample text input 3.}
    \label{fig:input_3_softpick_map}
\end{figure}

\newpage
\begin{figure}
    \centering
    \includegraphics[width=0.8\linewidth]{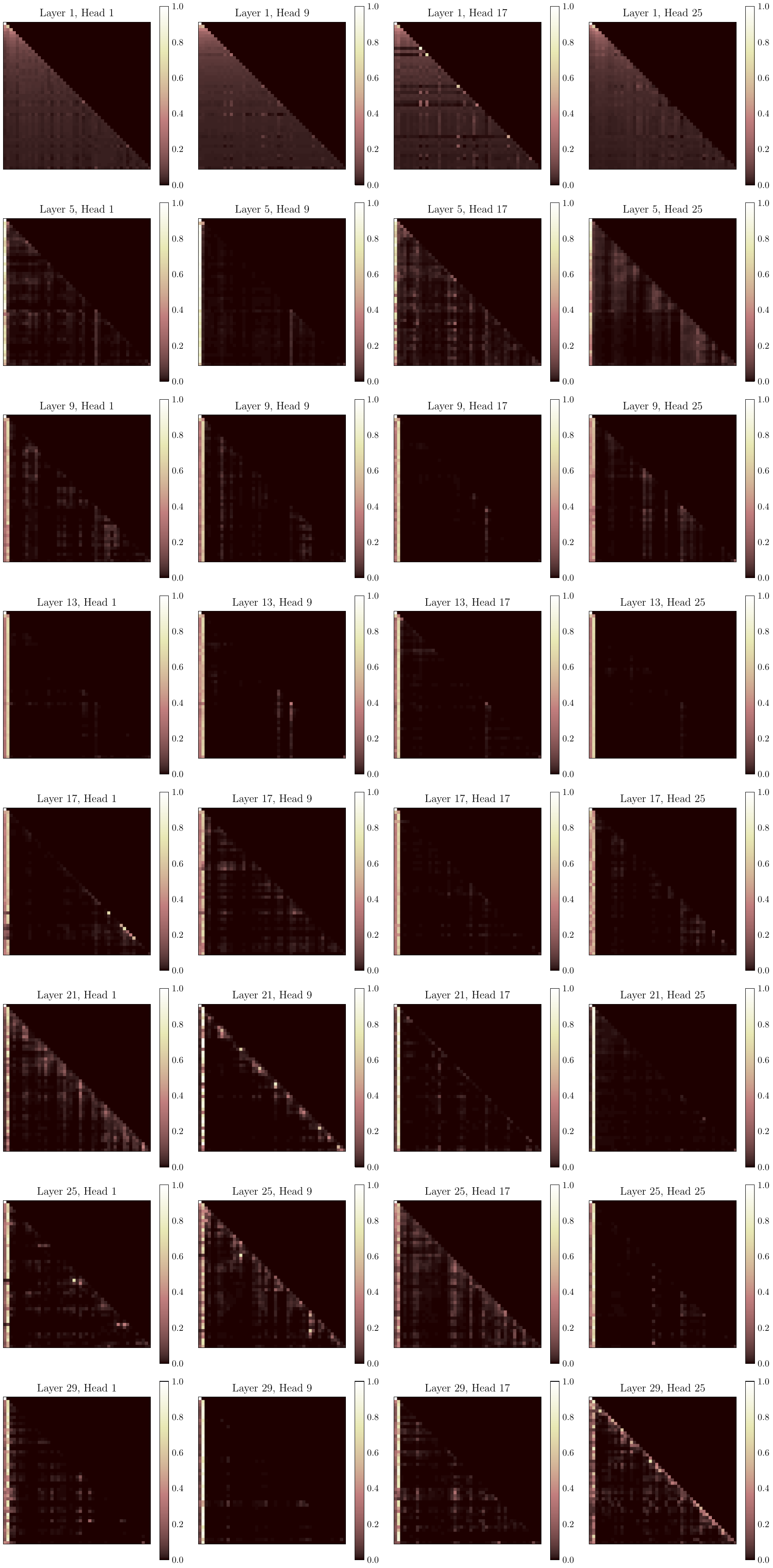}
    \caption{More attention maps of the softmax 1.8B model on sample text input 1.}
    \label{fig:1B_input_1_softmax_map}
\end{figure}

\newpage
\begin{figure}
    \centering
    \includegraphics[width=0.8\linewidth]{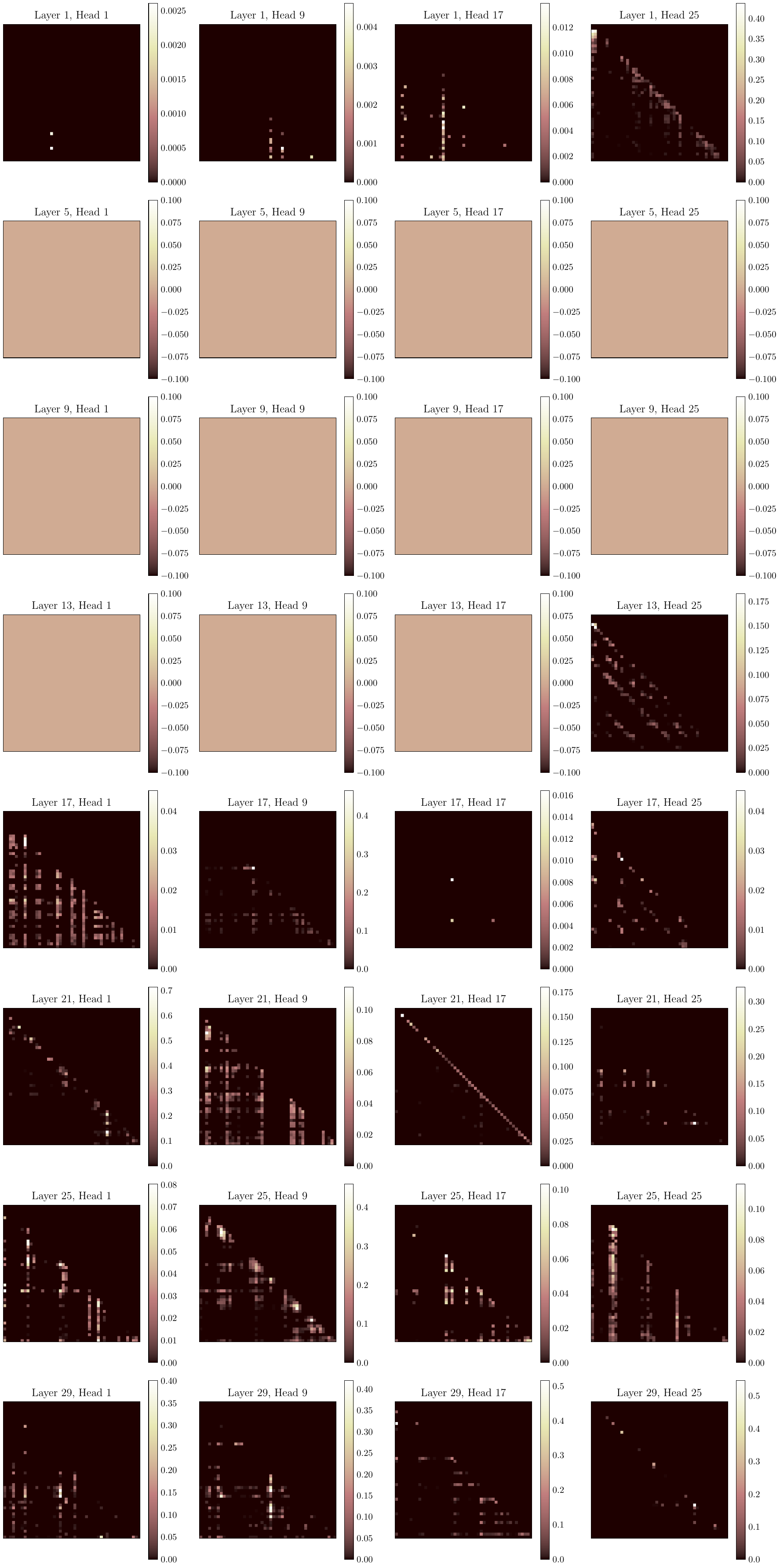}
    \caption{More attention maps of the softpick 1.8B model on sample text input 1.}
    \label{fig:1B_input_1_softpick_map}
\end{figure}

\newpage
\begin{figure}
    \centering
    \includegraphics[width=0.8\linewidth]{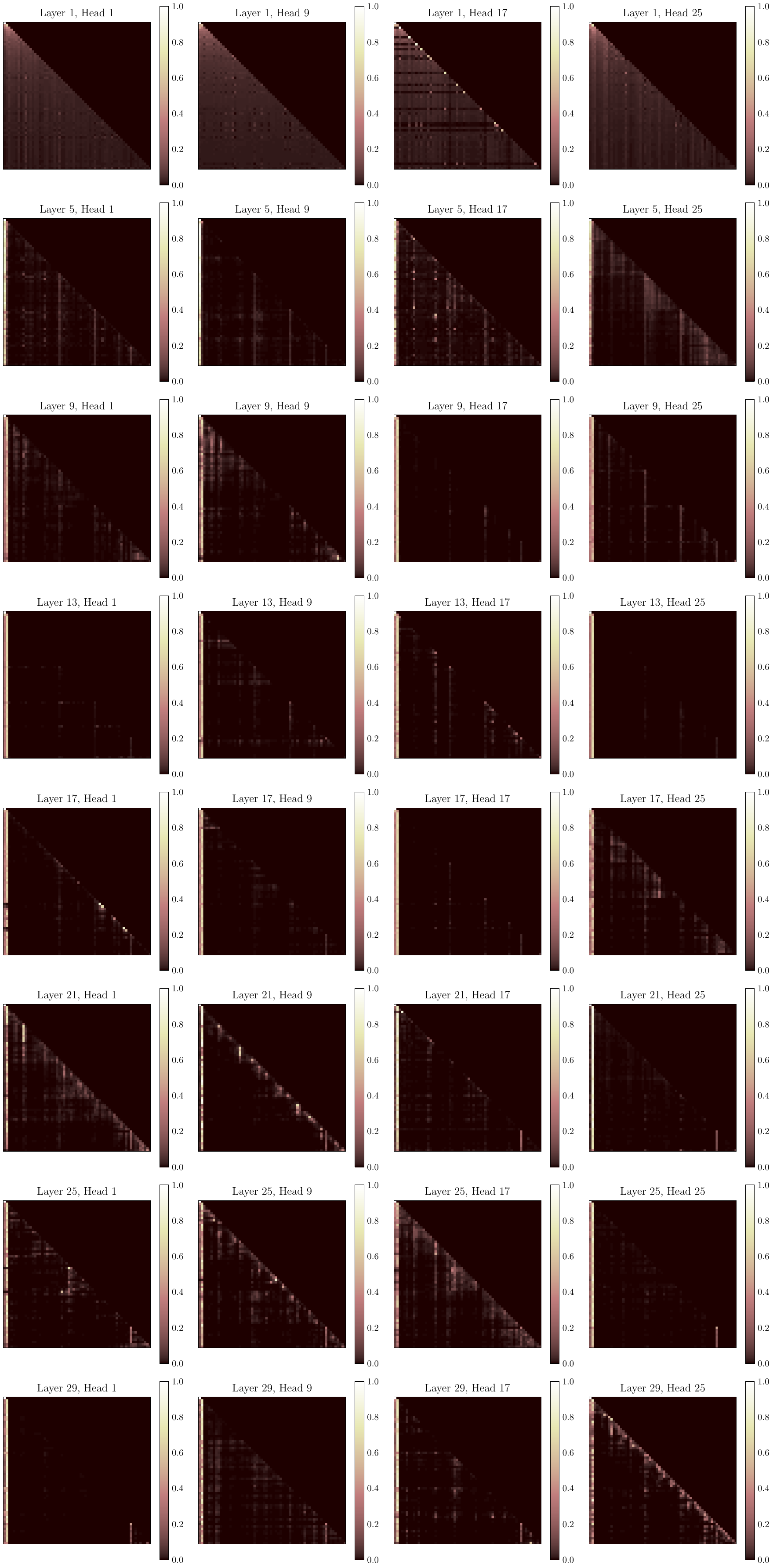}
    \caption{More attention maps of the softmax 1.8B model on sample text input 2.}
    \label{fig:1B_input_2_softmax_map}
\end{figure}

\newpage
\begin{figure}
    \centering
    \includegraphics[width=0.8\linewidth]{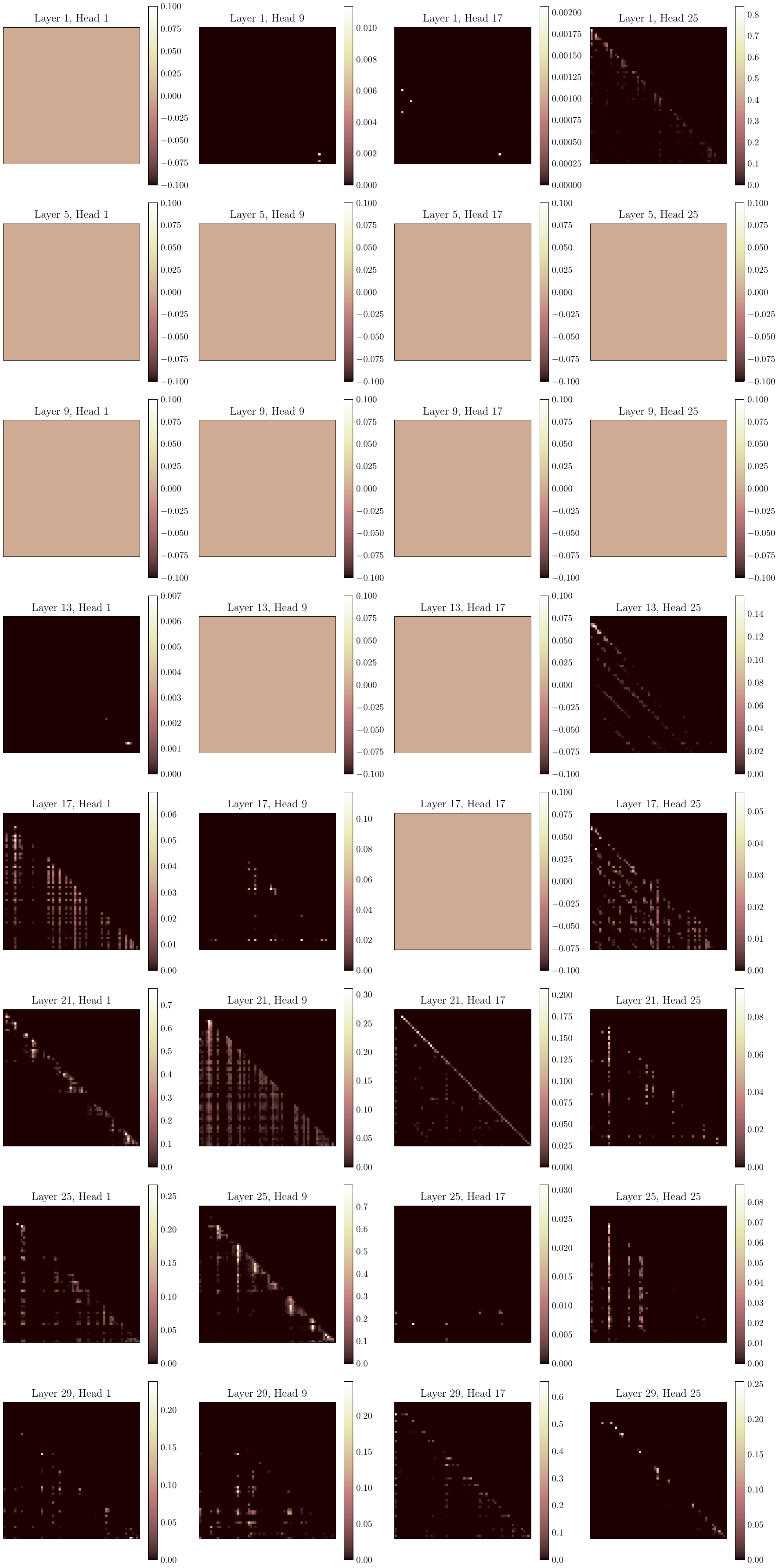}
    \caption{More attention maps of the softpick 1.8B model on sample text input 2.}
    \label{fig:1B_input_2_softpick_map}
\end{figure}

\newpage
\begin{figure}
    \centering
    \includegraphics[width=0.8\linewidth]{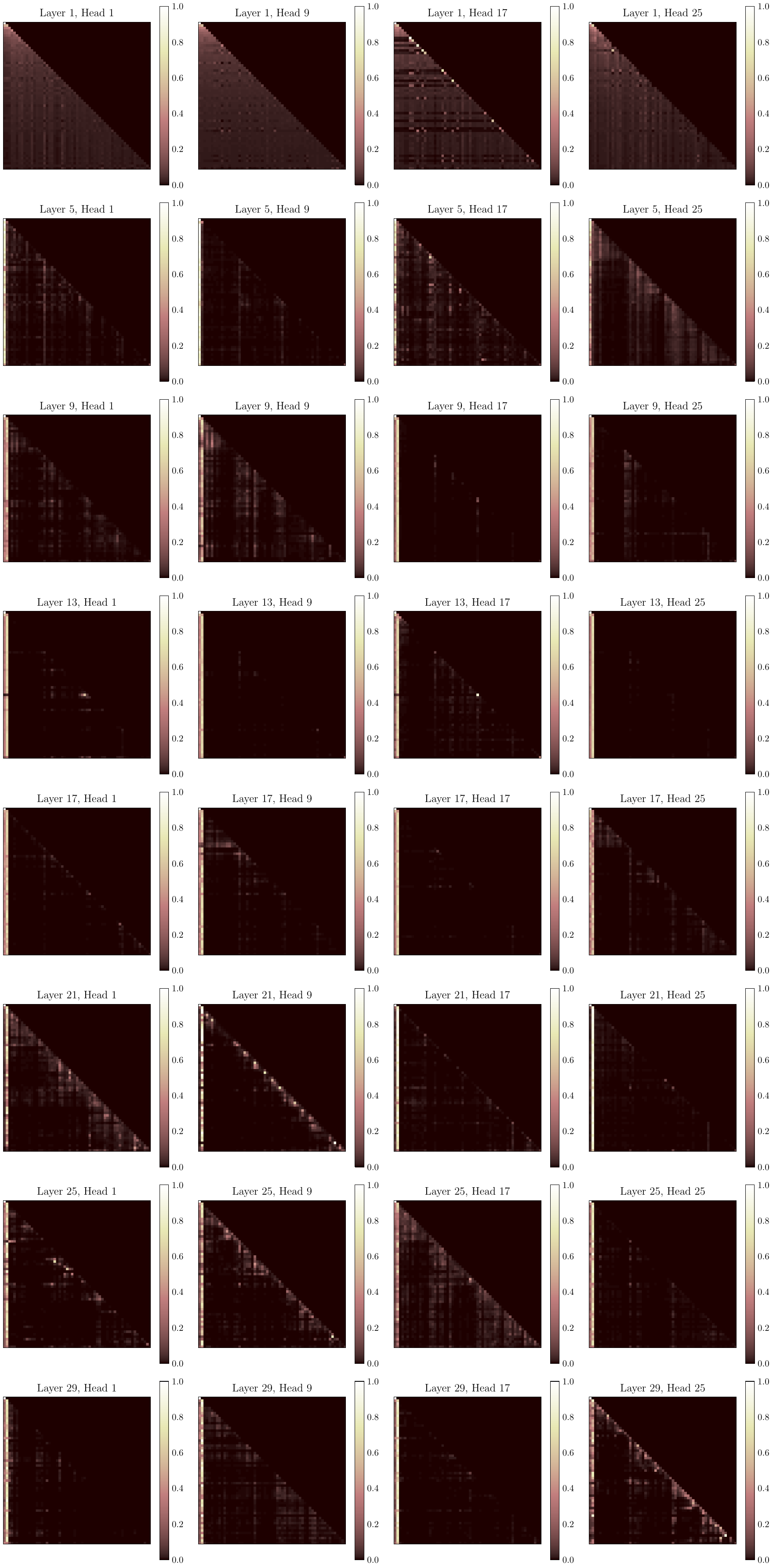}
    \caption{More attention maps of the softmax 1.8B model on sample text input 3.}
    \label{fig:1B_input_3_softmax_map}
\end{figure}

\newpage
\begin{figure}
    \centering
    \includegraphics[width=0.8\linewidth]{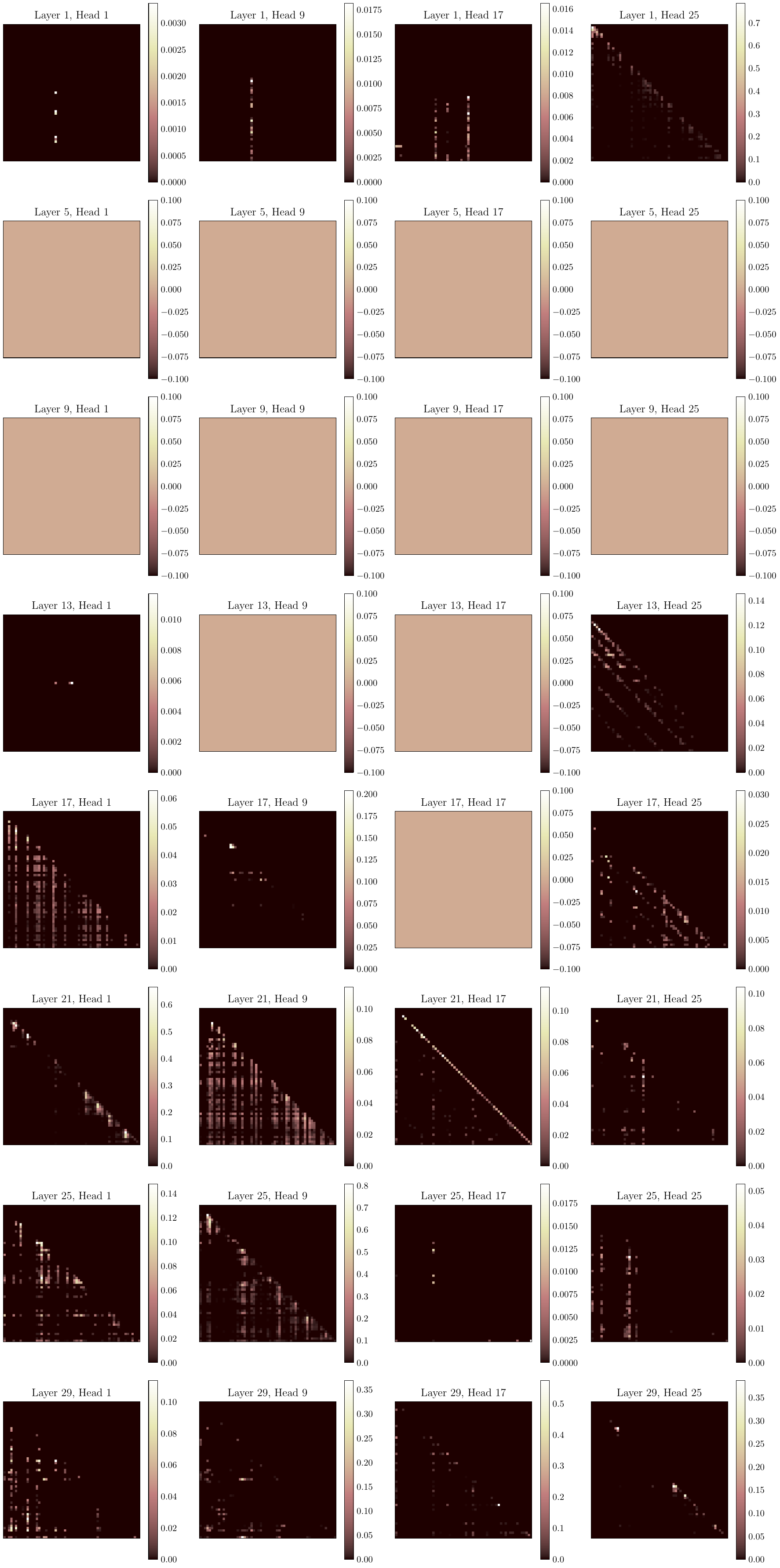}
    \caption{More attention maps of the softpick 1.8B model on sample text input 3.}
    \label{fig:1B_input_3_softpick_map}
\end{figure}

\end{document}